\documentclass[format=acmsmall, review=true]{acmart}

\renewcommand\footnotetextcopyrightpermission[1]{}
\newcommand\blfootnote[1]{%
  \begingroup
  \renewcommand\thefootnote{}\footnote{#1}%
  \addtocounter{footnote}{-1}%
  \endgroup
}
\usepackage{amsmath,amssymb,amsfonts}
\usepackage{booktabs}
\usepackage{multirow}
\usepackage{array}
\usepackage{pifont}
\usepackage{multirow}
\usepackage{subcaption}
\usepackage{enumitem}

\usepackage[linesnumbered,algoruled,boxed,lined]{algorithm2e}

\newcommand{\cmark}{\ding{51}}
\newcommand{\xmark}{\ding{55}}
\newcolumntype{C}[1]{>{\centering\let\newline\\\arraybackslash\hspace{0pt}}m{#1}}

\author{Nandan Kumar Jha}
\email{cs17mtech11010@iith.ac.in}
\orcid{0000-0001-6334-1740}
\affiliation{%
  \institution{Indian Institute of Technology Hyderabad}
  \streetaddress{Kandi}
  \city{Sangareddy}
  \state{Telangana}
  \postcode{502285}
  \country{India}
}

\author{Sparsh Mittal}
\email{sparshfec@iitr.ac.in}
\orcid{0000-0002-2908-993X}
\affiliation{%
  \institution{Indian Institute of Technology Roorkee}
  \streetaddress{}
  \city{Roorkee}
  \state{Uttarakhand}
  \postcode{247667}
  \country{India}
}

\author{Binod Kumar}
\email{binodkumar@ee.iitb.ac.in}
\affiliation{%
  \institution{Indian Institute of Technology Bombay}
  \streetaddress{Powai}
  \city{Mumbai}
  \state{Maharashtra}
  \postcode{400076}
  \country{India}
}

\author{Govardhan Mattela}
\email{cs18mds11034@iith.ac.in }
\affiliation{%
  \institution{Indian Institute of Technology Hyderabad}
  \streetaddress{Kandi}
  \city{Sangareddy}
  \state{Telangana}
  \postcode{502285}
  \country{India}
}

\begin{document}

\title{DeepPeep: Exploiting Design Ramifications to Decipher the Architecture of Compact DNNs}

\begin{abstract}
 
The remarkable predictive performance of deep neural networks (DNNs) has led to their adoption in service domains of unprecedented scale and scope. However, the widespread adoption and growing commercialization of DNNs have underscored the importance of intellectual property (IP) protection. Devising techniques to ensure IP protection has become necessary due to the increasing trend of outsourcing the DNN computations on the untrusted accelerators in cloud-based services. The design methodologies and hyper-parameters of DNNs are crucial information, and leaking them may cause massive economic loss to the organization. Furthermore, the knowledge of DNN's architecture can increase the success probability of an adversarial attack where an adversary perturbs the inputs and alter the prediction. 

In this work, we devise a two-stage attack methodology ``DeepPeep'' which exploits the distinctive characteristics of design methodologies to reverse-engineer the architecture of building blocks in compact DNNs. We show the efficacy of ``DeepPeep'' on  P100 and  P4000 GPUs. Additionally, we propose intelligent design maneuvering strategies for thwarting IP theft through the DeepPeep attack and proposed ``Secure MobileNet-V1''. {\em Interestingly}, compared to vanilla MobileNet-V1, secure MobileNet-V1 provides a significant reduction in inference latency ($\approx$60\%) and improvement in predictive performance ($\approx$2\%)  with very-low memory and computation overheads.
\end{abstract}



\maketitle
\thispagestyle{empty}

\section{Introduction} \label{sec:introduction}

Deep neural networks (DNNs) provide\blfootnote{Support for this work was provided by Semiconductor Research Corporation.} high accuracy on cognitive tasks, and hence, they are now spreading their wings to new mission-critical domains such as health care, defense, financial sector,  unmanned aerial vehicles, and autonomous driving.  The global revenue generated from the deployment and commercialization of artificial intelligence is forecast to increase from \$9.5 billion in 2018 to \$118.6 billion by 2025 \cite{AImarketForecast}. This makes DNN models a crucial intellectual property (IP) for the service providers. The IP of a DNN consists of the design methodologies employed in DNNs along with the architectural hyper-parameters (e.g., number of the filters in a layer, depth of the network, and filter size). A malicious entity who can access this information can counterfeit the IP. 
Further, the knowledge of DNN's architecture increases the success probability of adversarial attacks \cite{liu2016delving,2018_ICLR_joon_towards}. Several other attacks, such as membership inference attacks \cite{2018_Long_MembershipAttack,2017_Shokri_SP}, model extraction attack \cite{2016_Tram_SEC}, hyper-parameter stealing attack \cite{2018_Wang_SP}, and others,  also rely on the knowledge of DNN's architecture.  In other words, knowledge of DNN architecture allows an adversary to launch the attacks mentioned above. These attacks can have disastrous consequences,  especially in security-critical domains, such as banking and the stock market, and mission-critical domains, such as self-driving cars and robot-assisted surgery. {\em These trends and factors have underscored the importance of keeping the DNN model confidential}.

The training phase of computations in DNNs requires massive compute and memory resources  \cite{AIandCompute},  and hence,  DNN training is outsourced to the cloud services  \cite{kaplunovich2017cloud,mishra2010towards,del2018introducing}.  Further, to cope with the uncertainties and dynamic nature of the real-world environment, the cognitive applications deployed at the edge, such as self-driving cars and drones, share a massive amount of data with the cloud and frequently update the deployed models and policies \cite{2015_kehoe}. However, the low compute and storage capability, slower hardware/software update cycles, and extreme heterogeneity in edge devices make them unsuitable for continual and shared learning  \cite{stoica2017berkeley}. These factors necessitate the DNN computation on public clouds, which widen the exposure of DNNs to attacks.

GPUs have massive computing capability and are suitable for training DNNs on large datasets \cite{Russakovsky:2015:ILS:2846547.2846559}. Therefore, GPUs have been widely adopted by cloud service providers \cite{kaplunovich2017cloud,mishra2010towards} for DNN computations on the public clouds. 
Also, desktop-scale GPUs and mobile GPUs have been used for inference on autonomous vehicles.
Unlike trusted execution environments such as Intel's ``software guard extensions'' (SGX) \cite{costan2016intel}, ARM's TrustZone \cite{Zhao:2014:PRT:2666141.2666145} and Sanctum \cite{Costan2016SanctumMH}, GPUs do not provide hardware/software isolation and hence, do not guarantee a trusted execution environment \cite{mittal2019SurveySecurityGPU}. However, the aforementioned trusted execution environments suffer from performance degradation compared to the untrusted accelerators such as GPU \cite{tramer2018slalom}.  Therefore,  outsourcing DNN's computation on untrusted accelerators in public clouds increases the attack surface. {\em Hence, there is a crucial need to study DNN design methodologies and its robustness to side-channel attacks for preventing DNN IP theft}.

Recent research trends in DNN design have mainly focused on four aspects of the network, viz., {\em depth} \cite{He_2016_CVPR}, {\em width} \cite{BMVC2016_Zagoruyko}, {\em cardinality} \cite{Xie2017AggregatedRT}, and {\em attention} \cite{2018_Wang_Attention}. These aspects of network design have been employed for either (1) improving the representational power of the network with additional computation and parameter overhead, or (2) reducing the computation and parameter overheads without sacrificing the predictive performance of the network. However, the security aspect of network design, i.e., the robustness of design methodologies from IP counterfeiting standpoint, has been ignored.

In this paper, we ask a crucial question: {\em how vulnerable are the design methodologies of DNNs to side-channel attacks, and what is the minimum information adversaries need for counterfeiting the design IP}? To answer this question, we systematically and meticulously review the architecture of prominent compact DNNs and investigate the feasibility of deciphering the backbone architecture of compact DNNs through side-channels attacks. Through our thorough analysis, we found that the compact DNNs \cite{Howard2017MobileNetsEC, 8578572,Zhang2018ShuffleNetAE,Ma_2018_ECCV,Iandola2016SqueezeNetAA,Gholami2018SqueezeNextHN} are designed by repeatedly connecting few architectural building blocks which makes their design regular and modular and also reduces the number of design hyper-parameters. However, due to this, compact DNNs exhibit distinguishing properties in terms of the memory footprint, inference time, energy consumption, and resource utilization. These unique properties can easily be exploited by adversaries to infer the architecture of compact DNNs. Thus, we show that the existing approaches to designing compact DNNs have crucial security vulnerabilities.  {\em To the best of our knowledge, ours is the first work that investigates the security aspect of designing {\bf compact} DNNs from the IP theft standpoint}. 
We summarize our contributions as follows.

\begin{itemize}
\item We analyze the architectural implications of designing compact DNNs. This analysis reveals distinguishing characteristics of different design methodologies, which can be used by an adversary to decipher the employed design methodologies such as building blocks in compact DNNs (Section \ref{sec:SecurityVulnerability}).

\item We explore security aspects of designing compact DNNs and investigate the robustness of state-of-the-art design methodologies against design IP theft using various side-channel attacks (Section \ref{sec:SecurityVulnerability} and Section \ref{sec:SideChannelAttacks}).

\item We propose a two-stage black-box attack methodology termed ``DeepPeep'', (Figure \ref{fig:DeepPeepFlowChart}) which can be used by an adversary to decipher the architectural building blocks of a compact DNN based on its distinctive characteristics (Section \ref{sec:AttackModel}). We present experimental results of ``DeepPeep'' on state-of-the-art compact DNNs on two GPUs (P100 and P4000).
\begin{figure}[htbp]
\includegraphics[scale=0.3]{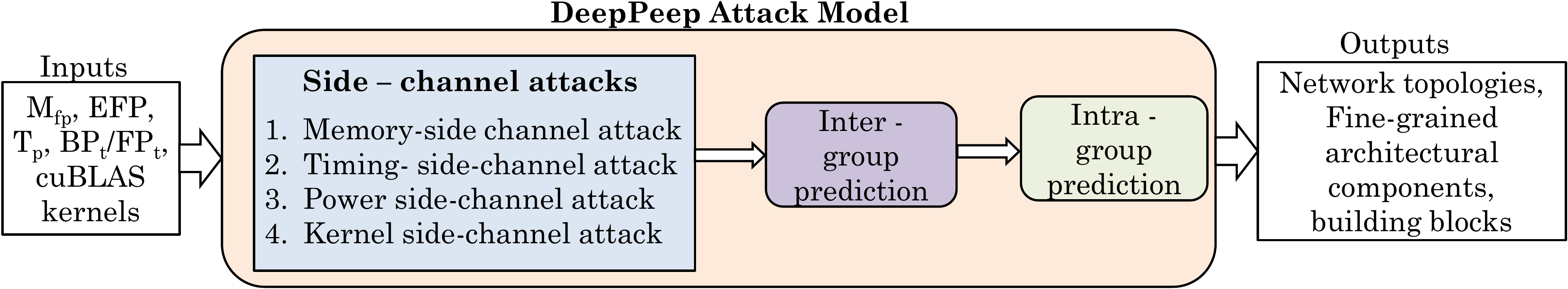} 
\caption{DeepPeep attack model}
\label{fig:DeepPeepFlowChart}
\end{figure}

\item We propose a  ``secure MobileNet-V1'' architecture, which is robust against the attack ``DeepPeep'' and prevents the IP theft. Compared to baseline MobileNet-V1, secure MobileNet-V1 achieves a 60.65\% reduction in inference time and, {\em most importantly}, there is a $\approx$2\% improvement in top-1 accuracy on both the ImageNet-1K and CIFAR-100  image classification datasets. We also provide generic design recommendations for security-aware DNN design in Section \ref{sec:DefenceMechanism}. 

\item We explain the generality and applicability of our proposed attack methodology ``DeepPeep'' on different model architectures, hardware platforms, and datasets (Section \ref{sec:GeneralityOfModel}).
\end{itemize} 

The rest of the paper is organized as follows. In section \ref{sec:background}, we define performance metrics and other terminologies used in this paper and provide a detailed explanation for experimental setup and methodology. In section \ref{sec:SecurityVulnerability}, we describe the building blocks in state-of-the-art compact DNNs along with their distinctive properties in terms of the percentage of cuBLAS kernels. We explore the various avenues for side-channel attacks to gain insights into the architecture of compact DNNs in section \ref{sec:SideChannelAttacks}. We describe our proposed black-box attack mechanism {\em DeepPeep} in section \ref{sec:AttackModel}. In section \ref{sec:DefenceMechanism}, we present the defense mechanism to thwart the side-channel attacks and provide recommendations for security-aware DNN design. The generality and applicability of ``DeepPeep'' on different DNN architectures, hardware platforms, and datasets have been investigated in section \ref{sec:GeneralityOfModel}. We discuss the machine-learning-based approach for deciphering the architecture of DNN and their demerits in section \ref{sec:Discussion}. In section \ref{sec:RelatedWork}, we present the related works on DNN security. Finally, we conclude with future work in section \ref{sec:conclusion}.

\section{Preliminaries} \label{sec:background}
Table \ref{tab:SymbolTable} shows the symbols used in this paper. Here, FPS is frames per second, ifmaps is input feature maps, ofmap is output feature maps, and GMACs is $10^9\times\#MACs$.

\begin{table} [htbp] 
\caption{Symbols used in this paper}
\label{tab:SymbolTable}\centering
\begin{tabular}{ |c|c|c|c| } \hline
 \textbf{Quantity (symbol)} & \textbf{Unit} & \textbf{Quantity (symbol)} & \textbf{Unit}  \\ \hline 
  \#MACs ($M_c$) & Millions & Memory-footprint ($M_{fp}$)   & MiB  \\
   \#Parameters ($P$) & Millions & Forward-pass time ($FP_t$)  & milliseconds  \\
    \#Activations ($A$) & Millions & Backward-pass time ($BP_t$)  & milliseconds  \\
    Energy per frame (EPF) & millijoule & Model-size ($M_s$) & MiB \\
      Batch size ($B$) & - & Energy efficiency ($E_e$)  & GMACs/Joule  \\ 
      Throughput($T_p$) & FPS &width/height of ifmaps ($S_{M}$)   & -   \\
      \#ifmaps ($M$) & - & width/height of ofmaps ($S_{N}$) & -  \\
      \#ofmaps ($N$)  & - & width/height of filter ($S_{F}$) & -  \\
   \hline
\end{tabular} 
\end{table}

\subsection{Metrics and Terminologies}

The metrics of interest are computed, as shown in the following equations. For these computations, we assume (1) spatial dimensions (height and width) of ifmaps are equal (2) spatial dimensions of ofmaps are the same, and (3) spatial dimensions of filters are equal. Note that the number of channels in the filter is equal to $M$, and the number of filters is equal to $N$.

\begin{align} 
\label{eqn:param}
\text {\# Parameters } &= N\times M\times S_{F}^2 \\
\label{eqn:act}
\text {\# Activations} &= N\times S_{N}^2 \\
\label{eqn:MACs}
\text {\# MACs} &= N\times M\times S_{N}^2\times S_{F}^2 \\
\label{eqn:energy}
\text{Energy per frame} &= (\text {Average power})\times (\text {$FP_t$})  \\
\label{eqn:EnergyEfficinecy}
\text {Energy efficiency } &= \frac{\text {Performance}}{\text {Power}} \nonumber \\
   &= \frac{\text{$B$}\times \text{$M_c$}}{\text{Energy per frame}}  
\end{align}

\textbf{Memory-footprint:} It depends on the maximum concurrent data at runtime, which includes (1) the total number of parameters, (2) concurrent activations and (3) gradients corresponding to activations and parameters. In other words, the memory-footprint depends on both the number of parameters and the number of activations. By comparison, model size depends only on the number of parameters in DNN. The correlation between the number of parameters and the number of activations entirely depends on the architecture of DNNs \cite{2019_Jha_VLSID}. That is, there is no direct correlation between model size and memory-footprint. 

\begin{figure}[htbp]
\begin{center}
\fbox{\includegraphics[scale=0.6]{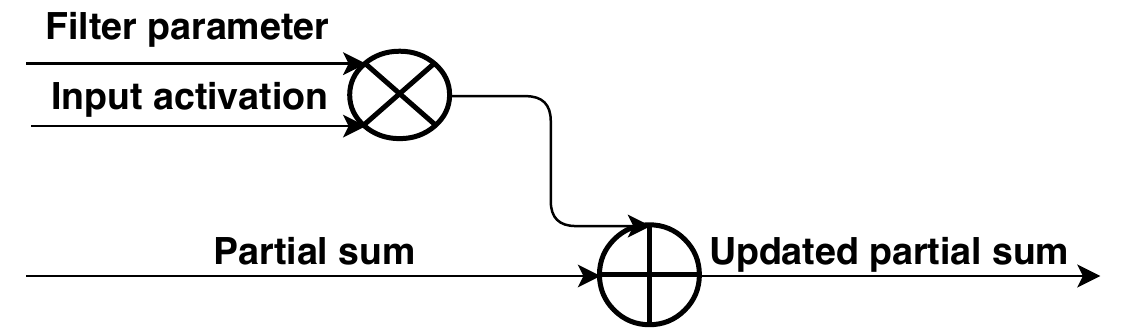}}
\caption{Illustration of a MAC operation. }
\label{fig:MAC}
\end{center}
\end{figure}

\textbf{Degree of data reuse:} It is defined as the number of arithmetic-operations per unit of operand fetched from different levels in memory-hierarchy. 
In convolutional layers, the arithmetic operations (MACs) have two types of operands, viz., filter-parameters, and input activations. This operation is shown in Figure \ref{fig:MAC}.  Hence, we express the degree of data reuse in terms of weight reuse ($\frac{M_c}{W}$) and activation reuse ($\frac{M_c}{A}$). Note that the partial sum (Figure \ref{fig:MAC}) is not a different type of operand; instead, it is expressed in terms of filter-weights and input-activations. 

\textbf{Energy per frame:} $EPF$ an amount of energy consumed in one forward pass through the entire network with a fixed spatial size image (Eqn. \ref{eqn:energy}). It depends on both the number of MACs operation and the energy-efficiency of DNN. Since energy-efficiency depends on the degree of data reuse, DNN with lower computational complexity, and poor data reuse exhibits higher $EPF$. Note that the number of MAC operations depends on the number of layers, shape/size of each layer, and types of convolution employed in the network. Thus,  $EPF$ varies quite significantly across different DNNs.

\subsection{Experimental Setup } \label{sec:ExperimentalSetup}
For our experiments, we use Caffe framework \cite{Jia:2014:CCA:2647868.2654889} and run the DNNs on Tesla P100-PCIE-12GB and Quadro P4000 GPUs. The former is a high-end data center scale GPU, and the latter is a desktop GPU. Hence, these GPUs have significantly different compute, bandwidth, and memory resources (Table \ref{tab:GPUsFeatures}). Since GPUs have an ample amount of computing capability and memory resources, performing the experiments on GPUs allows analyzing the characteristics of DNN's design methodologies without getting bottlenecked by system resources. This capability is especially useful at larger batch sizes. 

\begin{table} [htbp]
  \caption{Configuration of GPUs used in our experiments}
   \label{tab:GPUsFeatures} 
   \resizebox{0.95\textwidth}{!}{
  \begin{tabular}{ c|c|c|c|c|c|c } 
    \toprule
    
 \textbf{GPU}   & \textbf{\# SMs} & \textbf{\# core} & {\bf Global memory (MB)}& \textbf{Peak bandwidth (GB/s) }  & \textbf{Peak $T_p$ (TFLOPS)} & \textbf{TDP (watts)}   \\
      \midrule
   P100 & 56 & 3584 & 12193 & 549  & 9.3  & 250    \\
    \midrule
       P4000 & 14 & 1792 & 8118 &243  & 5.2  & 105    \\ 
       
    \bottomrule
  \end{tabular}}
\end{table}

{\bf Power measurement on GPU:} We use an inbuilt \textit{nvidia-smi} utility to measure the power on Nvidia's GPUs. Since \textit{nvidia-smi} is a high-level utility and sampling rate depends on that of the inbuilt power sensor in GPU \cite{bridges2016understanding}, accurately measuring the power consumption on GPU is a non-trivial task and requires extra care. The inbuilt sensor's sampling rate is not only quite low but also GPU specific, i.e., varies from GPU to GPU. For instance, on P100 and P4000 GPUs, the sampling rate is $\approx50Hz$ and $\approx1Hz$, respectively. To circumvent the challenge mentioned above,  we run DNNs for a higher number of iterations (forward pass and backward pass simultaneously) and get the stabilized power readings. More precisely, at batch size one, we run 500 iterations for DNNs with lower inference time (e.g., SqueezeNet and GoogLeNet), and 100 iterations for the DNNs with higher inference time. Note that at larger batch sizes, we run the same number of iterations for all the DNNs (in Table \ref{tab:CDNNblocks}). Similar to \cite{wu2017squeezedet}, we sample the readings of \textit{nvidia-smi} tool at every 0.1 ms, and once the power reading becomes stable, we take the average power (reported by \textit{nvidia-smi}) as the power consumption of a DNN. 

Similarly, as above, we take an average over a large number of iterations for forward and backward propagation time ($FP_t$ and $BP_t$ respectively) while running DNN in Caffe \cite{Jia:2014:CCA:2647868.2654889} deep learning framework. This way, we increase the robustness and mitigate the effect of noise on the readings. Memory-footprint {$M_{fp}$} is measured as the global memory occupied in GPU during the DNN computation in Caffe.  Further, to know the distinctive properties of different building blocks in compact DNNs and gain more insights about the architectural ingredients of basic building blocks in all the compact DNNs, we have conducted experiments with various batch sizes ($B$).  We take $B$ as either the power of two or the integer multiples of the number of streaming processors (SMs) in GPUs. For example, on P100 GPU, we take $B$ as 1, 4, 8, 16, 32, and 56, whereas on P4000 GPU, we take $B$ as 1, 4, 8, 14, and 28 \footnote{On GPUs P100 and P4000, MobileNet-V2 and DenseNet are not able to run  with higher $B$ (such as $B$ = 28,56) due to their high memory-footprint.}

\section{Distinctive Properties of Building Blocks in Compact DNNs} \label{sec:SecurityVulnerability}

In this section,  we first discuss the architectural building blocks in compact DNNs, such as fire module, inception module, dense block, along with their fine-grained architectural components, such as branching and residual/skip connections. (Table \ref{tab:CDNNblocks}). This analysis of architectural ingredients in compact DNNs would help in reverse engineer the basic building blocks in compact DNNs through the side-channel attacks. Then, we show the GPU kernel (cuBLAS) profiling results for all the compact DNNs listed in Table \ref{tab:CDNNblocks}. We then analyze the properties of these cuBLAS kernels, which would help decipher the fine-grained architectural components present in building blocks of compact DNNs.

\subsection{Architectural Building Blocks in Compact DNNs} \label{sec:SecurityBuildingBlocks}

Most compact DNN designs employ a single building block in a repeated manner to reduce the number of design hyper-parameter choices and make design optimization easier. Also, it makes DNN's design more regular and modular. These basic building blocks in compact DNNs can be broadly categorized as fire module, depthwise convolution (DWConv), channel shuffling, dense block, and inception module (Table \ref{tab:CDNNblocks}). 
For example, SqueezeNet and SqueezeNext use the fire module, whereas MobileNet and ShuffleNet DNNs use DWConv. 

\begin{table} [htbp]
\caption{DNN groups with similar architectural building blocks. In each group, the DNN shown in bold is the reference DNN for our experiments.}
\label{tab:CDNNblocks}
\centering
\resizebox{1.0\textwidth}{!}{
\begin{tabular}{ |c|c|C{1.3cm}|c|C{1.5cm}|C{1.1cm}|C{1.6cm}||c|c|C{2.5cm}|C{3cm}|} 
 \hline
 \textbf{DNN group}&\textbf{Model Name}  &\textbf{Fire module} & \textbf{DWConv} & \textbf{Channel shuffling} & \textbf{Dense block} & \textbf{Inception module} & \textbf{PWConv} &\textbf{Branching} & \textbf{Residual/skip connections} & \textbf{Asymmetric filter decomposition  } \\
 \hline
  \multirow{2}{*}{SqueezeNet} &{\bf SqueezeNet-V1.0 } \cite{Iandola2016SqueezeNetAA}	  & \cmark &\xmark&\xmark&\xmark&\xmark &\cmark&\cmark&\xmark&\xmark\\
   & SqueezeNet-V1.1 \cite{Iandola2016SqueezeNetAA}   &\cmark&\xmark&\xmark&\xmark&\xmark &\cmark&\cmark&\xmark&\xmark\\
   \hline
  \multirow{5}{*}{SqueezeNext} &1.0-G-SqNxt-23	\cite{Gholami2018SqueezeNextHN}  &\cmark&\xmark&\xmark&\xmark&\xmark  &\cmark&\cmark&\cmark&\cmark\\
   &{\bf 1.0-SqNxt-23}  \cite{Gholami2018SqueezeNextHN}	  &\cmark&\xmark&\xmark&\xmark&\xmark  &\cmark&\cmark&\cmark&\cmark\\
   &1.0-SqNxt-23v5	\cite{Gholami2018SqueezeNextHN}  &\cmark&\xmark&\xmark&\xmark&\xmark  &\cmark&\cmark&\cmark&\cmark\\
   &2.0-SqNxt-23  	\cite{Gholami2018SqueezeNextHN}  &\cmark&\xmark&\xmark&\xmark&\xmark   &\cmark&\cmark&\cmark&\cmark\\
   &2.0-SqNxt-23v5	 \cite{Gholami2018SqueezeNextHN} &\cmark&\xmark&\xmark&\xmark&\xmark   &\cmark&\cmark&\cmark&\cmark\\
   \hline
  \multirow{2}{*}{MobileNet} &{\bf MobileNet-V1} \cite{Howard2017MobileNetsEC}	      &\xmark&\cmark&\xmark&\xmark&\xmark   &\cmark&\xmark&\xmark&\xmark\\
   &MobileNet-V2	 \cite{8578572}      &\xmark&\cmark&\xmark&\xmark&\xmark  &\cmark&\xmark&\cmark&\xmark\\
   \hline
   \multirow{2}{*}{ShuffleNet}&{\bf ShuffleNet-V1} \cite{Zhang2018ShuffleNetAE}    &\xmark&\cmark&\cmark&\xmark&\xmark &\cmark&\cmark&\cmark&\xmark\\
   &ShuffleNet-V2  \cite{Ma_2018_ECCV}   &\xmark&\cmark&\cmark&\xmark&\xmark &\cmark&\cmark&\cmark&\xmark\\
   \hline
  DenseNet &{\bf DenseNet-121}	\cite{8099726}      &\xmark&\xmark&\xmark&\cmark&\xmark &\cmark &\xmark &\cmark  &\xmark\\
   \hline
  \multirow{3}{*}{InceptionNet} &{\bf GoogLeNet}	\cite{Szegedy_2015_CVPR}      &\xmark&\xmark&\xmark&\xmark&\cmark &\cmark&\cmark&\xmark&\xmark\\
   &Inception-V2	\cite{Szegedy_2016_CVPR}       &\xmark&\xmark&\xmark&\xmark&\cmark &\cmark&\cmark&\xmark&\cmark\\
   &SE-BN-Inception \cite{Hu2018SqueezeandExcitationN}	  &\xmark&\xmark&\xmark&\xmark&\cmark &\cmark&\cmark&\cmark&\cmark\\
   \hline 
\end{tabular}}
\end{table}

The building-blocks mentioned above in compact DNNs are made up of fine-grained architectural ingredients illustrated in the last four columns of  Table \ref{tab:CDNNblocks}. For example, dense blocks have skip connections, whereas the fire module in SqueezeNet has pointwise convolution (PWConv) and branching. {\em We notice that building blocks in compact DNNs share standard fine-grained architectural components, but they are arranged in a different manner}. Thus, our intuition is, these basic blocks would exhibit the distinctive characteristics which can be exploited by an adversary to reverse engineer the architecture of these basic blocks.

\subsection{Analysis of GPU Profiling Results} \label{sec:GPUprofilingResults}
To understand the effect of architectural building blocks on GPU's compute and memory resource utilization, we perform kernel-level profiling using {\tt nvprof} on P100 GPU.  We analyze three cuBLAS kernels {\tt Gemmk1}, {\tt Gemv2T} and {\tt Gemv2N} because these kernels are present in all the compact DNNs listed in Table \ref{tab:CDNNblocks}, except SqueezeNet variants. The percentage contribution of the aforementioned cuBLAS kernels in one iteration of forward and backward pass (successively) is shown in Table \ref{tab:KernelResults}.

\begin{table} [htbp]
\caption{The contribution (in \%) of cuBLAS kernels in one iteration of forward  and backward pass successively. }
\label{tab:KernelResults}
\centering
\begin{tabular}{ |c|c|c|c|} 
 \hline
 \textbf{Model Name} & \textbf{Gemv2T($\%$)} &\textbf{Gemv2N($\%$)}& \textbf{Gemmk1($\%$)} \\
 
    \hline
    SqueezeNet-V1.0   &   0    & 0     & 0             \\
    
    SqueezeNet-V1.1   &   0    & 0    &  0  \\
   \hline
    1.0-G-SqNxt-23    &  34.53 & 5.33  & 9.13           \\
    
    1.0-SqNxt-23      &  36.53 & 5.64   & 9.66           \\
    
    1.0-SqNxt-23v5    &  27.78 & 6.35  & 10.85 \\
    
    2.0-SqNxt-23      &  30.65 & 4.75  & 8.25           \\
    
    2.0-SqNxt-23v5    &  21.49 & 4.94  & 8.35          \\
    \hline
     MobileNet-V1     &  \textbf{59.23} & \textbf{30.55}  & 0.63 \\
     
     MobileNet-V2     &  \textbf{60.31} & \textbf{28.79}  & 0.80 \\
     \hline
     ShuffleNet-V1  &  {\bf 45.37}   & {\bf 29.50}   &  4.39          \\
      
     ShuffleNet-V2 &  {\bf 43.81}   & {\bf 30.58}   & 3.39         \\
     \hline
    DenseNet-121      &  18.19 &  3.66   & 7.32           \\
    \hline
    GoogLeNet         &  0.18  & 0.18  & 0.05           \\
    
    Inception-V2      &  5.12  & 0.03  & 3.69           \\
    SE-BN-Inception   &  5.75  & 0.03  & 3.35    \\
    \hline
\end{tabular}
\end{table}

\begin{figure}[htbp]  
\begin{center}
\begin{subfigure}{0.45\textwidth}
\includegraphics[scale=0.25]{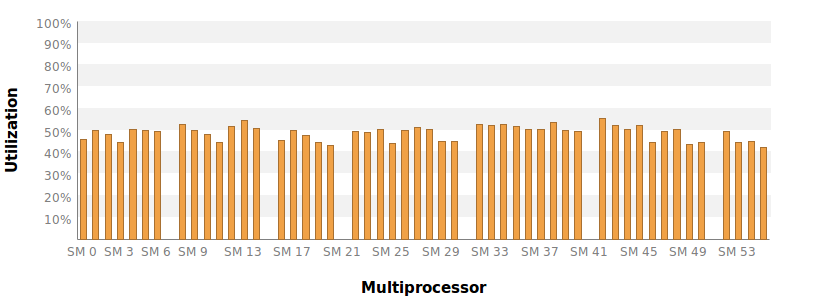} 
\caption{Gemmk1} \label{fig:2a}
\end{subfigure}
\begin{subfigure}{0.45\textwidth}
\includegraphics[scale=0.25]{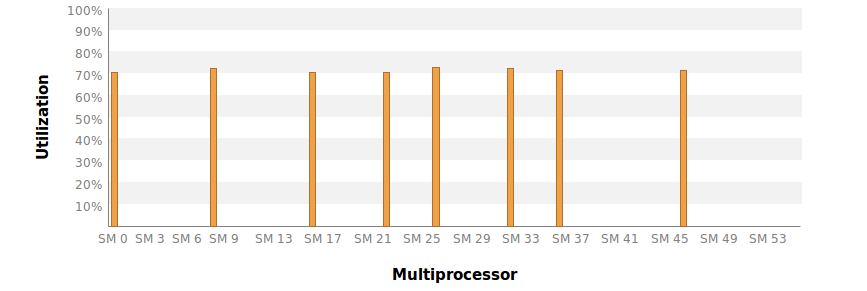} 
\caption{Gemv2T} \label{fig:2b} 
\end{subfigure}
\begin{subfigure}{0.5\textwidth}
\includegraphics[scale=0.30]{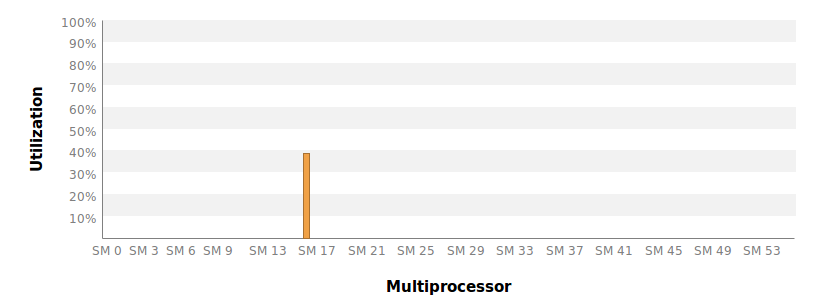}
\caption{Gemv2N} \label{fig:2c}
\end{subfigure}

\caption{Utilization of SMs on P100 GPU by cuBLAS kernels {\tt Gemmk1}, {\tt Gemv2T}, and {\tt Gemv2N}} \label{fig:Kernelutilization}
\end{center}
\end{figure}

We use Nvidia visual profiler to analyze the performance bottleneck and resource utilization by individual cuBLAS kernels. The contribution of memory dependency and instruction dependency in the total stall in cuBLAS kernels are shown in Table \ref{tab:KernelPropertiesAnalysis}. Also, the SM utilization during the execution of cuBLAS kernels is shown in Figure \ref{fig:Kernelutilization}.

{\bf Stalls in cuBLAS kernels:}
As shown in Table \ref{tab:KernelPropertiesAnalysis}, the contribution of memory dependency in total stalls in  {\tt Gemmk1} and {\tt Gemv2T} kernels are  54.4\% and 44.3\% respectively and hence, both these kernels are memory bound. Similarly,  56.9\% stalls in kernel {\tt Gemv2N} are due to instruction dependency, and hence, {\tt Gemv2N} is instruction bound. The performance of memory-bound kernels can be boosted by increasing the data-level parallelism, such as increasing the $B$. Whereas, the performance of instruction-bound kernel can be increased only by introducing more instruction-level parallelism. Hence, unlike {\tt Gemv2N}, the poor SM utilization of {\tt Gemv2T} kernel can be improved with larger $B$. In other words, the  SM utilization of DNNs with a higher percentage of {\tt Gemv2T} kernel can be improved using a larger $B$. By contrast, SM utilization of DNNs with a higher percentage of {\tt Gemv2N} kernel cannot be improved even at a higher $B$.

\begin{table}[htbp]
\caption{Analysis of kernel properties}
\label{tab:KernelPropertiesAnalysis}
\centering
\begin{tabular}{ |c|c|c|c| } 
 \hline
 \textbf{Attributes/stall reasons}&  \textbf{Gemmk1} & \textbf{Gemv2T} &\textbf{Gemv2N}\\
 \hline
 Memory dependency ($\%$) &54.4 & 44.3   & 3.6  \\
 SM utilization      & Excellent & Poor   & Very poor \\
 \hline
\end{tabular} 
\end{table}

{\bf Analysis:}
{\tt Gemm} (general matrix-matrix multiplication) is a dense matrix computation kernel and hence, it is {\em compute intensive and memory efficient} \cite{7995252}. By comparison,  {\tt Gemv} (general matrix-vector multiplication) is a sparse matrix computation kernel and hence, it is {\em bandwidth limited and memory inefficient} \cite{7995252}. In effect, {\tt Gemm} computation has better data reuse and also, it is  faster than  {\tt Gemv} computation \cite{Notallops}. The {\tt Gemmk1} kernel is a variant of {\tt Gemm} kernel, whereas {\tt Gemv2T} and {\tt Gemv2N} kernels are variant of {\tt Gemv} kernel.

As shown in Figure  \ref{fig:Kernelutilization}, the {\tt Gemmk1} kernel computations result in uniform and better SM utilization while {\tt Gemv2T} and {\tt Gemv2N} kernel  computations have very poor SM utilization. In fact, the poor data layout in {\tt Gemv2N} kernel \cite{cuBLAS_libraries} renders all the SM underutilized, except one. Therefore, DNNs with higher percentage of {\tt Gemmk1} kernel have better resource utilization compared to DNNs with higher percentage of {\tt Gemv2T} and {\tt Gemv2N} kernels. {\em This analysis of cuBLAS kernels and the understanding of their performance in terms of resource utilization facilitates ``timing side-channel attack''} (Section \ref{sec:SideChannelAttacks}).

\section{Side-Channel Attack Modeling} \label{sec:SideChannelAttacks}

In this section, we first present the threat model and our assumptions, for the attack on compact DNNs. We discuss the essence of predicting only the parametric layers in the building blocks of compact DNNs. 
Then, we describe the side-channel attacks used in ``DeepPeep'' attack model.

\begin{figure}[htbp] 
\centering
\includegraphics[scale=0.35]{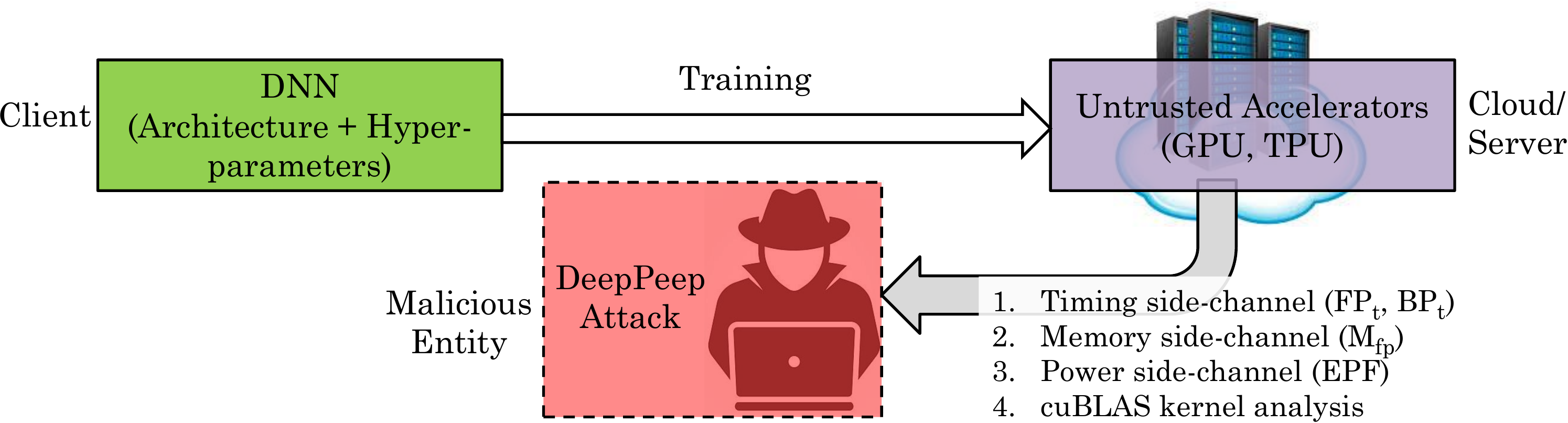} 
\caption{ Threat model: DNN training is outsourced to public cloud where malicious entities try to decipher the architecture of DNN through side-channel attacks.} 
\label{fig:ThreatModel}
\end{figure}

{\bf Threat Model and assumptions:} 
We assume that either the cloud service provider itself is malicious, or the third parties who are running their models on the same cloud are illegitimately trying to gain information about other models. We assume our attack methodology as a black-box setting where the structure and the internal states (such as $W$, $A$, and $M_c$) of DNNs running on the cloud are invisible to the adversary. The performance metrics which are known to adversary are listed in Table \ref{tab:AssumptionsTable}. Our threat model  is shown in Figure \ref{fig:ThreatModel}. 

\begin{table} [htbp] 
\caption{Assumption on adversary's knowledge}
\label{tab:AssumptionsTable}\centering
\resizebox{1.0\textwidth}{!}{ 
\begin{tabular}{ |c|c| } \hline
 \textbf{Known states/quantities } & \textbf{Unknown states/quantities} \\
 \hline 
 $FP_t$, $BP_t$, $M_{fp}$, power and energy consumption of DNNs, {\tt nvprof} output  & $W$, $A$, $M_c$, depth of DNNs, building blocks, (Table \ref{tab:CDNNblocks}). 
 \\ 
 \hline
\end{tabular} }
\end{table}

{\bf Validity of our assumptions:}
To meet the growing need for running DNN computations on public cloud, the cloud service providers use a distributed set of servers shared with other competitors where one cloud providers can not control others \cite{stoica2017berkeley}. The complexity of software stack is growing to allow the running applications to leverage tools such as DNN frameworks, data processing frameworks, and log ingestion.  Hence, a security hole in even one component of these complex software stacks or distributed execution environments can compromise the DNN application.  Furthermore, in contrast to the white-box attack where the target DNN model is visible to the adversary, the black-box attack is more realistic in a real-world scenario. However, to make a black-box attack feasible, the adversary will need to make more queries (than required in a white-box attack) through side-channel attacks.

{\bf Do we need to predict non-convolutional layers also?}
In our proposed attack methodology  ``DeepPeep'', we intend to predict the architecture of principle building blocks in terms of their architectural ingredients. We do not predict fully connected (FC) layers and other non-convolutional layers such as ReLU, pooling, batch normalization layers.  There are three reasons for this. First, except for FC, all layers are non-parametric, and there are no weights/parameters to be updated (learned) during the training of DNNs. That is, these layers do not encode any information about the inputs.  Second, the position of these layers in DNN is well-known and fixed in almost all state-of-the-art DNN. For instance, all the DNNs employ either a global average pooling layer or an FC layer as the last layer (before the softmax layer). Similarly, the position of other non-convolutional layers, between two consecutive parametric layers (convolutional/FC), is well-known. Third, unlike convolutional layers, these layers are not a determinant of internal representation in DNN. In summary, there is no IP associated with these layers, and hence, \textit{there is no need to predict the presence of non-convolutional layers}.

We now discuss the side-channel attack modeling in the proposed attack model ``DeepPeep''. These side-channel attacks exploit the distinctive characteristics of building blocks and other fine-grained architectural components. In other words, these attacks are based on our systematic study on (1) the effect of architectural components on performance metrics $M_{fp}$, $E_e$, $T_p$, and (2) the percentage of cuBLAS kernels in DNNs.

\subsection{Memory Side-channel Attack} 
This attack is based on the understanding of the effect of DNN's architectural components, such as basic building blocks and other fine-grained architectural ingredients, on total memory-footprint ($M_{fp}$). $M_{fp}$ has a strong dependence on $A$, which in turn depends on both the network topology (linear and non-linear) and type of operations (DWConv, PWConv, and others) \cite{Notallops,2019_Jha_VLSID}. In linear networks, such as AlexNet and VGGNet, layers are sequentially connected only to their immediate neighboring layers. By contrast, in non-linear networks, such as SqueezeNet and DenseNet, layers are connected to multiple adjacent layers \cite{wang2018superneurons}. Therefore, in DNNs with non-linear network topology, the memory-overprovisioning issue gets aggravated and results in higher $M_{fp}$ \cite{wang2018superneurons}. This issue becomes even more pronounced when (1) non-linear DNNs are deeper \cite{wang2018superneurons},  and (2) non-linear DNNs are executed at higher $B$ \cite{2019_Jha_VLSID}.

\begin{figure*}[htbp]  
\includegraphics[scale=0.252]{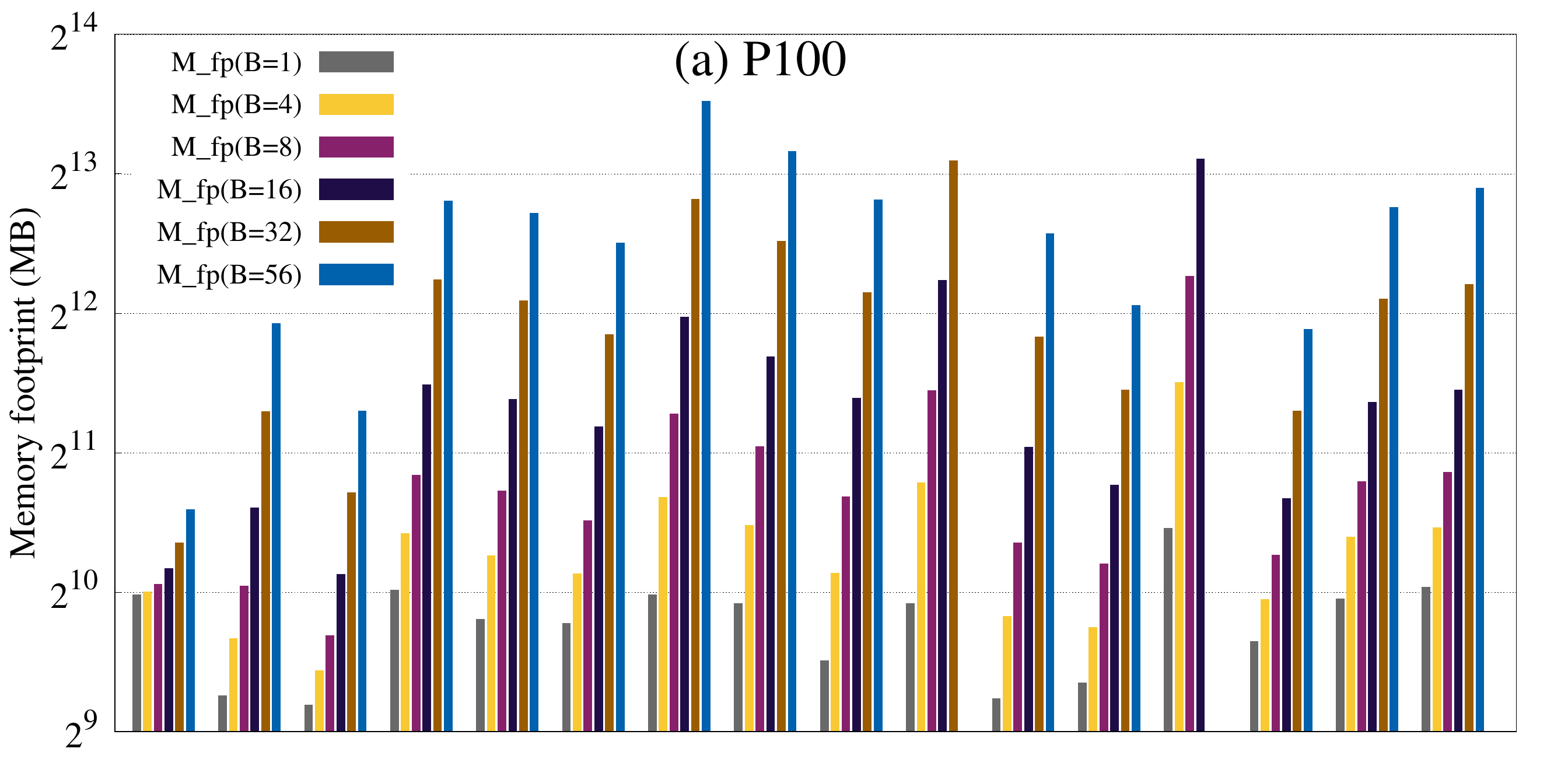}  
\includegraphics[scale=0.252]{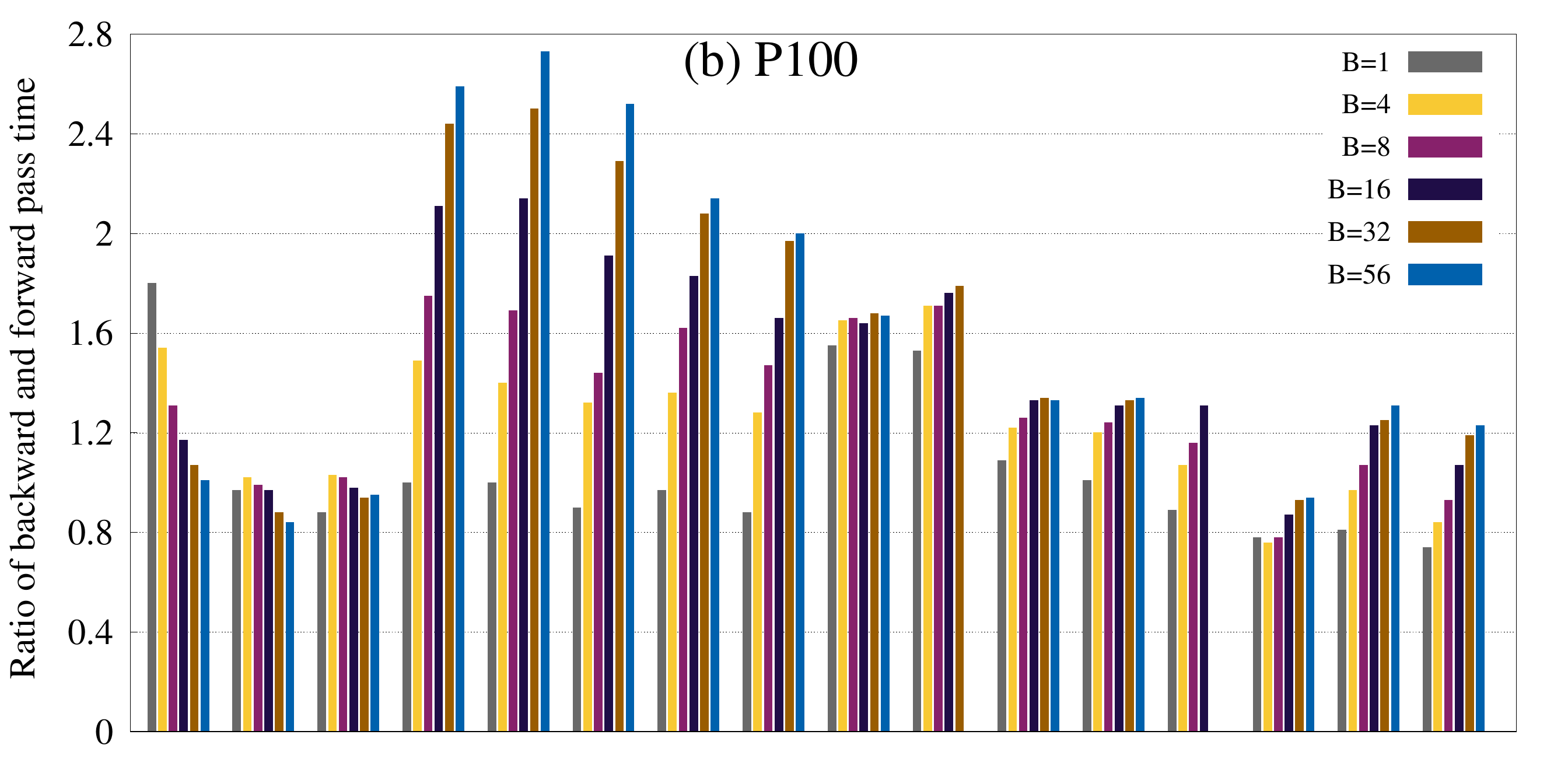} \\
\includegraphics[scale=0.252]{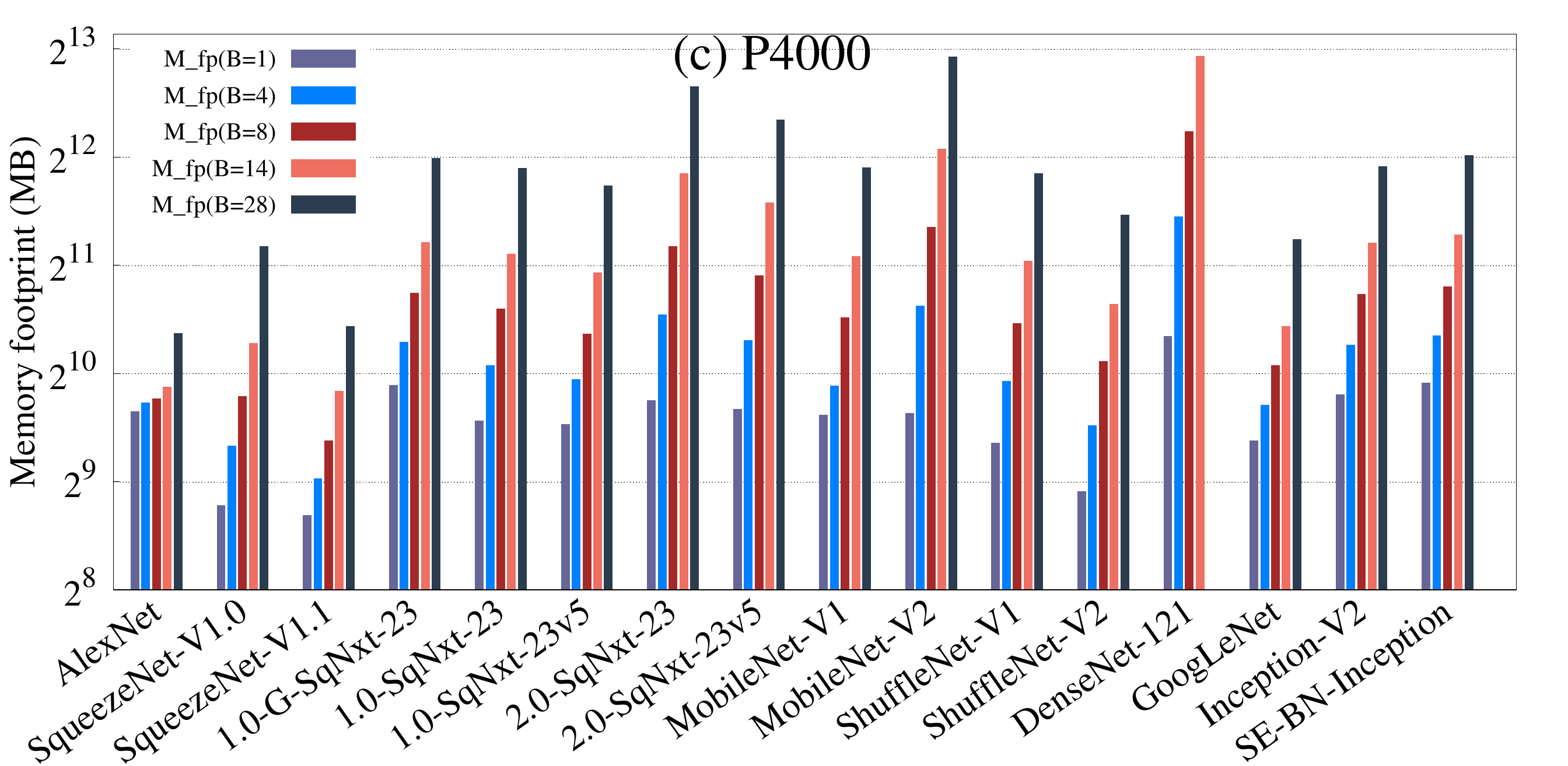} 
\includegraphics[scale=0.252]{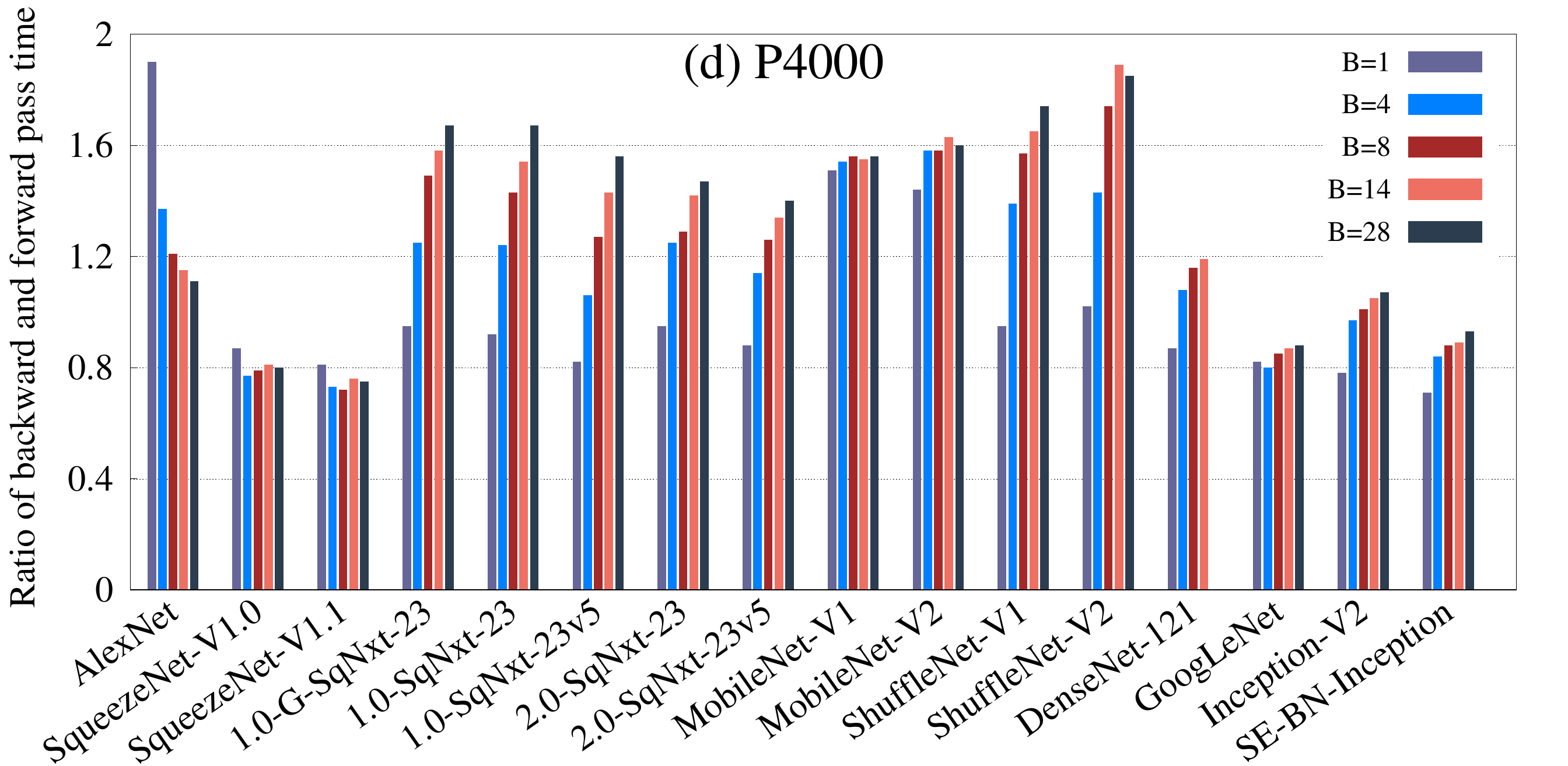} 

\caption{Memory-footprint and $\frac{BP_t}{FP_t}$ ratio on both P100 GPU (Figure (a) and (b) respectively) and on P4000 GPU (Figure (c) and (d) respectively) }
\label{fig:MemfpAndFPbpRatio}
\end{figure*}

Notice that SqueezeNext variants have very high $\frac{A}{P}$ ratio and 1.0-G-SqNxt-23 has higher $M_{fp}$ (Table \ref{tab:ResultsSummary}) than AlexNet even when the former has 112$\times$ fewer $P$ than the latter (Table \ref{tab:Modelattributes}) . The fire module in SqueezeNext has all the fine-grained architectural components listed in last four columns in Table \ref{tab:CDNNblocks} which renders $\frac{A}{P}$ ratio very high. 
As shown in  Figure \ref{fig:MemfpAndFPbpRatio}(a) and \ref{fig:MemfpAndFPbpRatio}(c), the increase in $M_{fp}$ with $B$ for all the compact DNNs  is substantially high compared to that in AlexNet. In fact, in DenseNet, $M_{fp}$ increases exponentially with $B$ because the number of skip connections in dense blocks vary exponentially with the number of layers  \cite{8099726}.

\begin{table} [htbp]
\caption{ All the values are measured on P100 GPU. $T_p$ is measured at the batch size shown in parenthesis. }
\label{tab:ResultsSummary}
\centering
\begin{tabular}{ |c|c|c|c|c|} 
 \hline
 \multirow{2}{*}{\textbf{Model Name}} & \textbf{$FP_t$}   & \textbf{$M_{fp}$}    &  \textbf{$E_e$}   &  $T_p$   \\
   &  ($B=1$) &  ($B=1$)  &   ($B=4$) &     \\
 \hline
     AlexNet        & 2.1     & 1015    & 75.94  & 4000  ($B$=56)   \\  
    \hline
    SqueezeNet-V1.0 & 3.8     & 615     & 40.95  & 1414  ($B$=56)       \\
    
    SqueezeNet-V1.1  & 3.5    & 587     & 27.07   & 2488   ($B$=56)     \\
   \hline
    1.0-G-SqNxt-23   & 24.4   & \textbf{1019}    & 1.69   & 665    ($B$=56)  \\
    
    1.0-SqNxt-23     & 24.5   & 885     & 2.16   & 691   ($B$=56)         \\
    
    1.0-SqNxt-23v5   & 23.8   & 867     & 1.92   & 856   ($B$=56) \\
    
    2.0-SqNxt-23     & 28.2   & 995    & 3.75   & 418   ($B$=56)     \\
    
    2.0-SqNxt-23v5    & 27.7  & 957     & 3.86   & 538   ($B$=56)           \\
    \hline
     MobileNet-V1& 29.4  & 733     & 1.65   & 42  ($B$=56)  \\
     MobileNet-V2& 42.2  & 970     & 0.80   & 28   ($B$=32)  \\
    \hline
    DenseNet-121      & 33.0  & \textbf{1405}   & 8.41  & 164   ($B$=16)         \\
    \hline
    GoogLeNet         & 11.2  & 801    & 21.95   & 1167   ($B$=56)          \\
    
    Inception-V2      & 19.0  & 987    & 14.05   & 602   ($B$=56)          \\
    \hline
\end{tabular}
\end{table}

 \begin{enumerate} []
 \item [] {\bf Observation 1:} \label{obs:memorySCA} The architectural components such as skip connections in dense blocks, residual connection, branching, etc. lead to non-linear topology in DNNs. They also increase  $\frac{A}{P}$ ratio which is greater than one for all the compact DNNs in Table \ref{tab:Modelattributes}.  
Further, the rate of increase in  $M_{fp}$ with $B$ depends on the rate of increase in $\frac{A}{P}$ ratio with the depth of network and $B$. {\em This correlation between the $M_{fp}$ and DNN's architecture can be exploited by an adversary to decipher the network topology employed in compact DNNs}.

\end{enumerate}

\subsection{Timing Side-channel Attack}
\begin{figure}[htbp]
\centering
\includegraphics[scale=0.252]{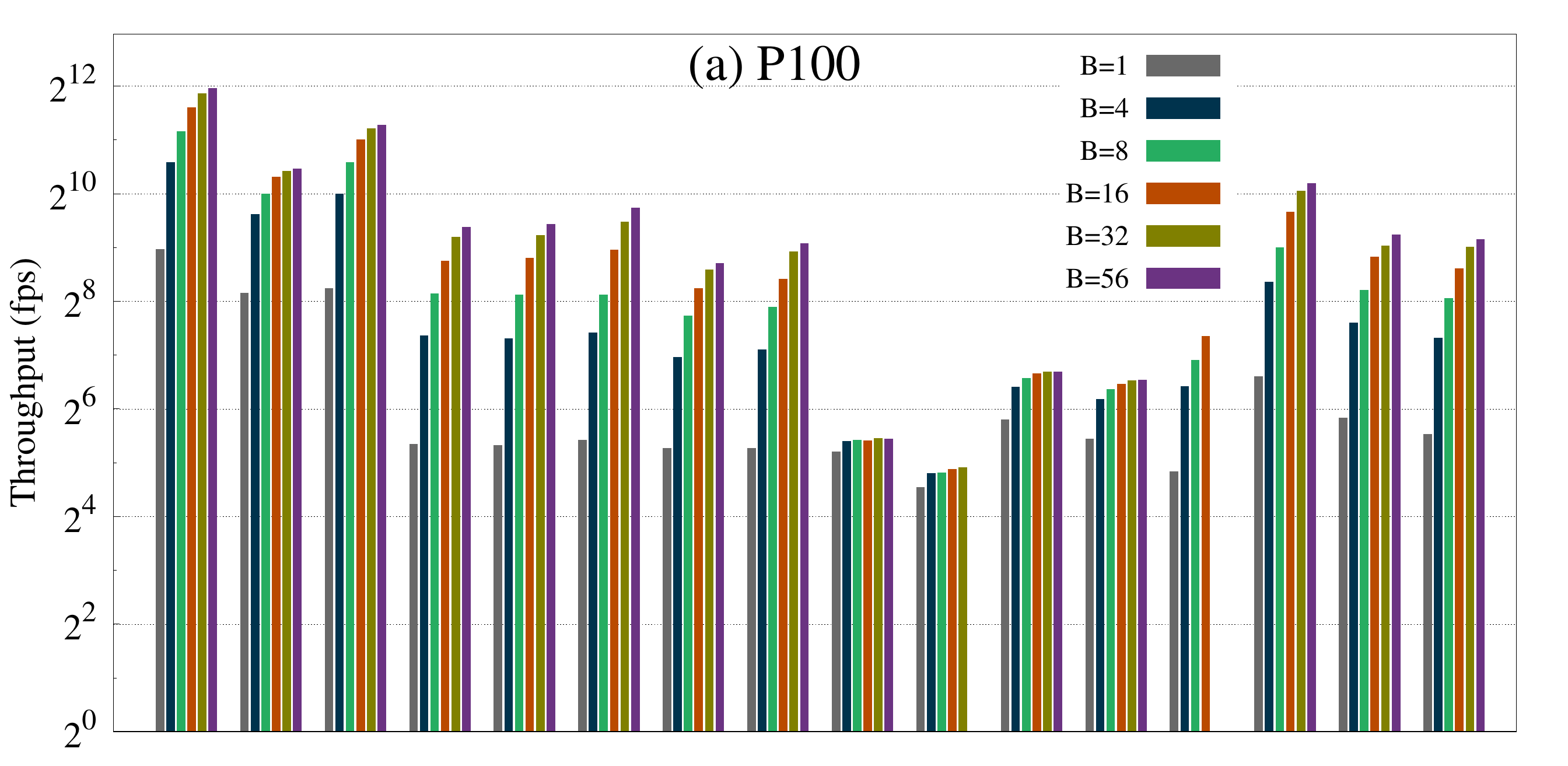}  
\includegraphics[scale=0.252]{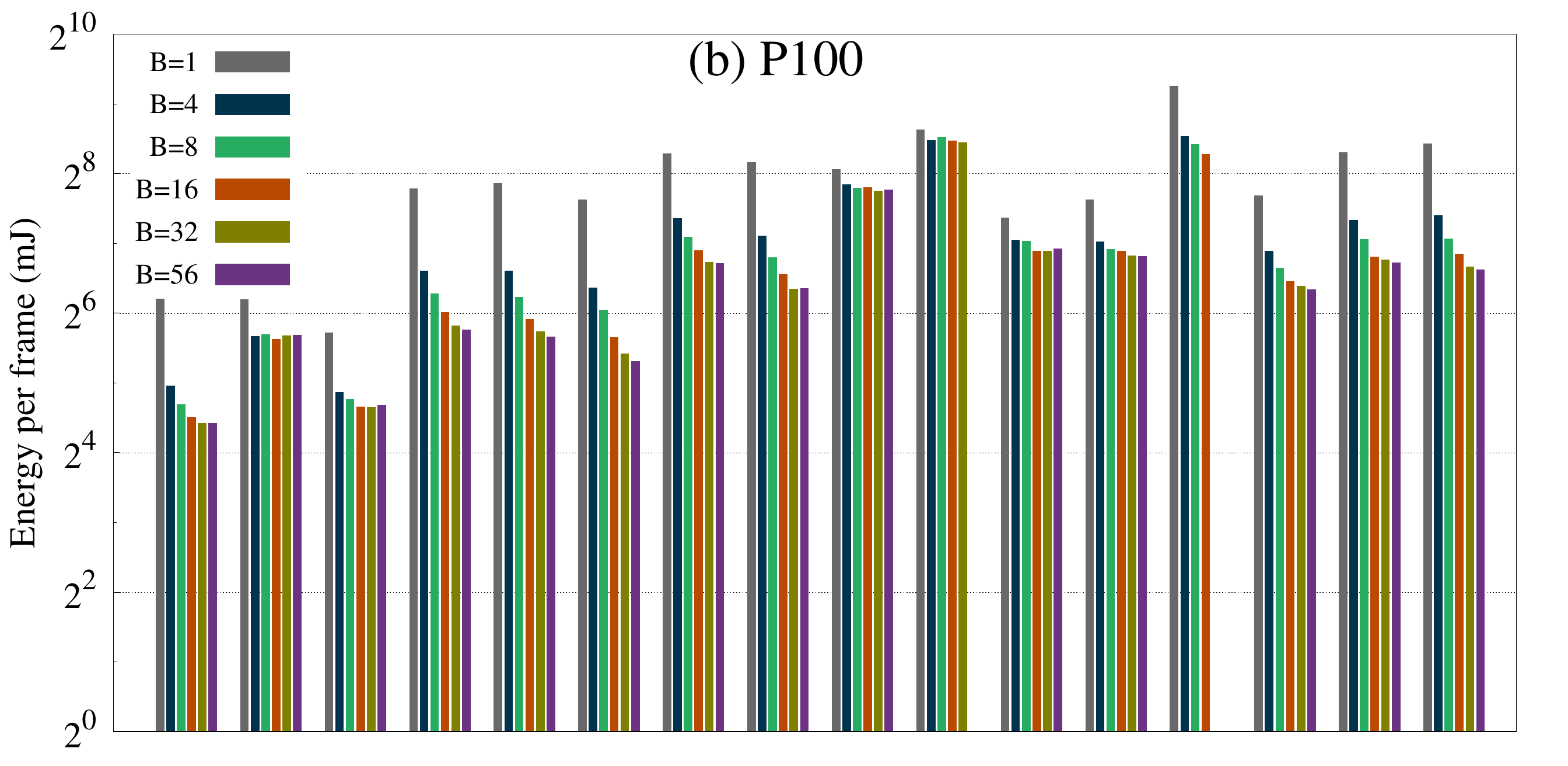} \\
\includegraphics[scale=0.252]{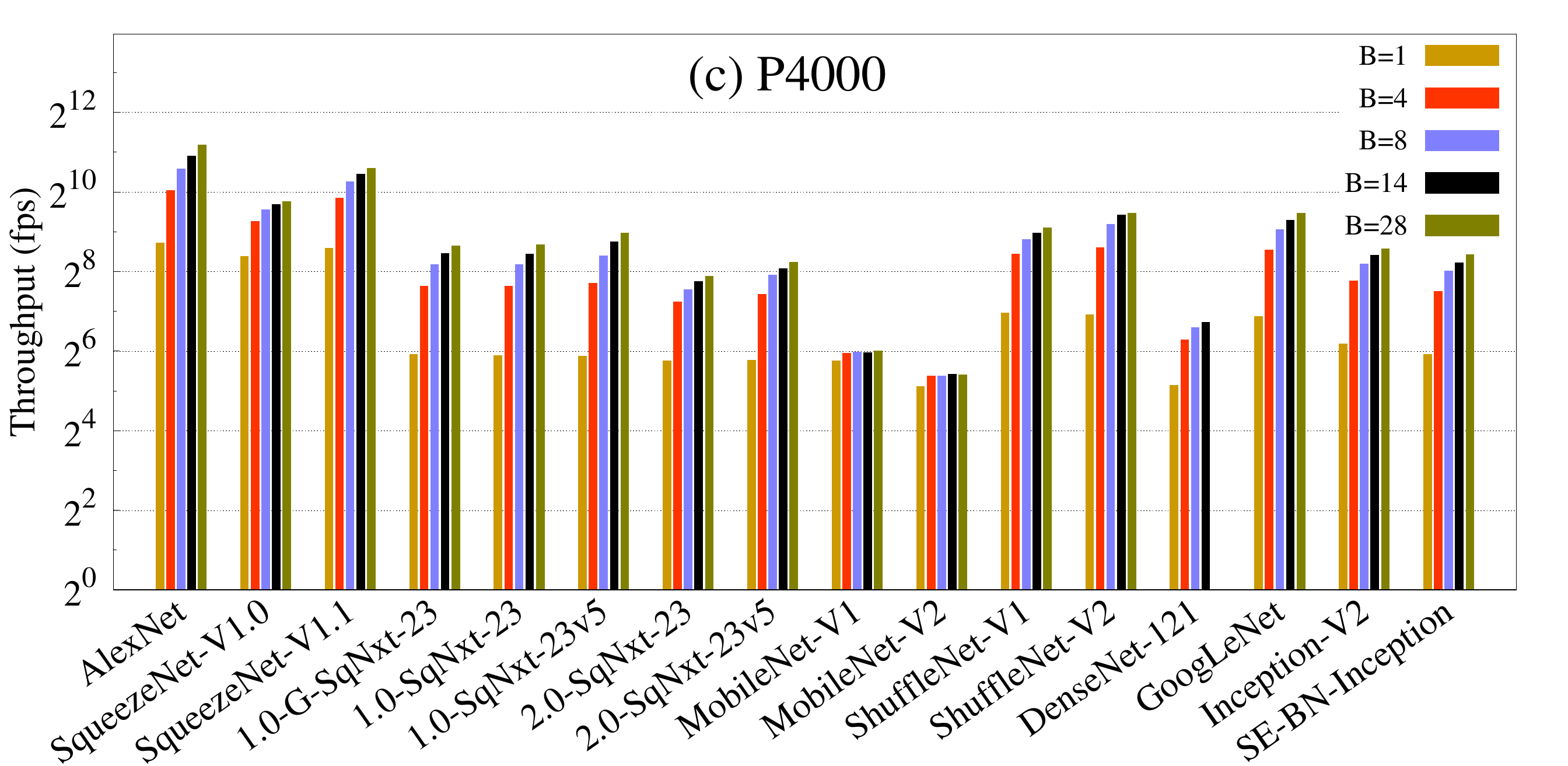}  
\includegraphics[scale=0.252]{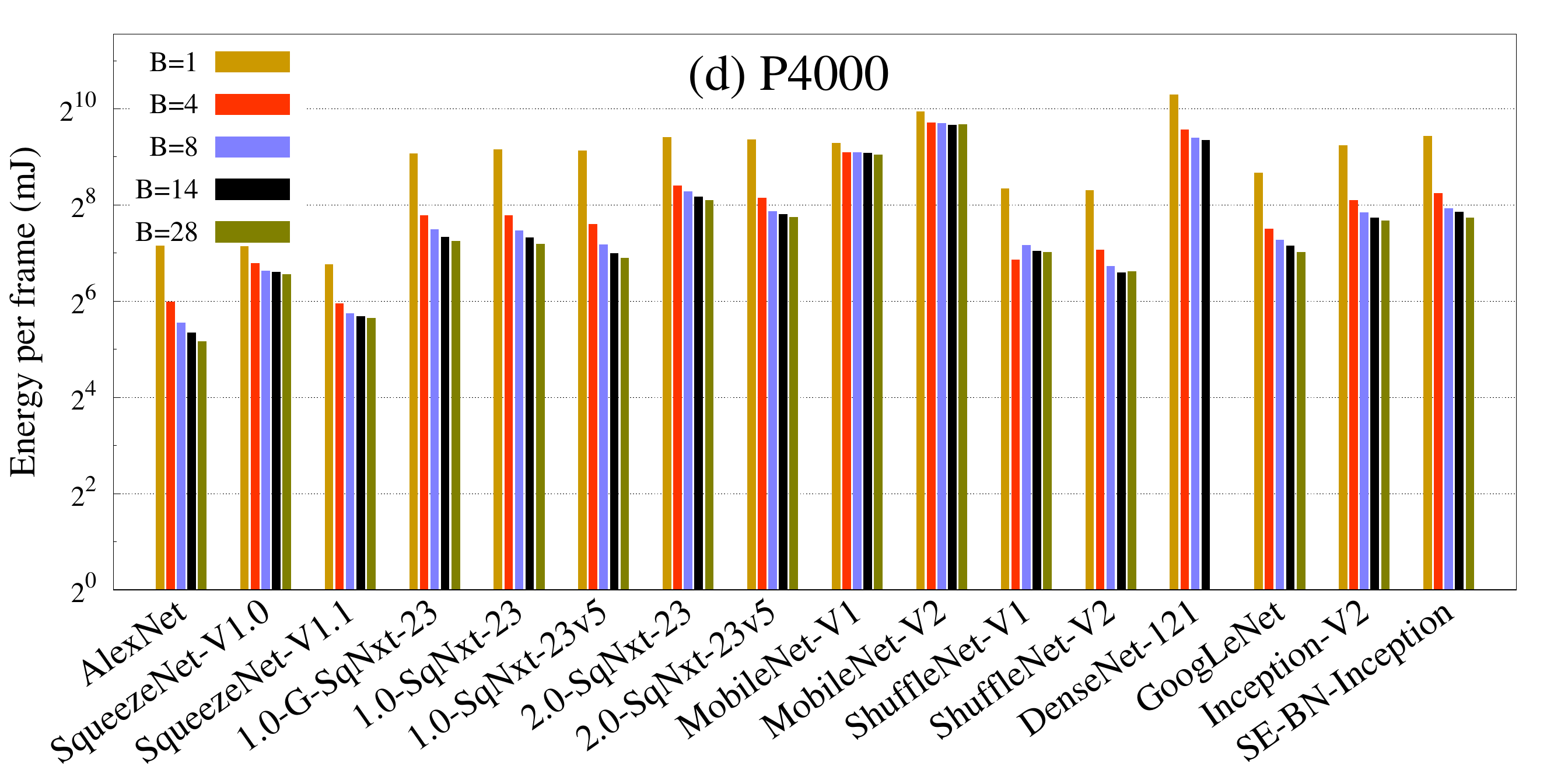} 

\caption{ Throughput ($T_p$) and energy per frame ($EPF$) on both P100 GPU (Figure (a) and (b) respectively) and on P4000 GPU (Figure (c) and (d) respectively) }
\label{fig:ThroughputEnergy}
\end{figure}

This attack is based on our thorough study of GPU compute resource (SM) utilization with increasing $B$.  
We measure  $\frac{BP_t}{FP_t}$ ratio  (Figure \ref{fig:MemfpAndFPbpRatio}(b) and \ref{fig:MemfpAndFPbpRatio}(d)) and $T_p$ (Figure \ref{fig:ThroughputEnergy}(a) and \ref{fig:ThroughputEnergy}(c)) for all the compact DNNs with increasing $B$.

\begin{enumerate}  []
\item [] {\bf Observation 2:}\label{obs:timingSCA1} In all the compact DNNs, $\frac{BP_t}{FP_t}$ ratio either remains constant or increases with $B$. By comparison,   in AlexNet, this ratio decreases   gradually with increasing $B$. {\em Through this timing side-channel attack, one can easily distinguish compact DNNs from large DNNs such as AlexNet}. 
\end{enumerate}

Among the compact DNNs, SqueezeNet variants (V1.0/V1.1), MobileNet variants (V1/V2), and GoogLeNet have constant $\frac{BP_t}{FP_t}$ ratio with increasing $B$. Notice that, in case of SqueezeNet variants and GoogLeNet, $\frac{BP_t}{FP_t}$ ratio is lower than one, however, MobileNet variants have $\frac{BP_t}{FP_t}$ ratio greater than one.

As discussed in Section \ref{sec:GPUprofilingResults}, higher percentage of {\tt Gemv2T} and {\tt Gemv2N} leads to poor SM utilization. However, the latter has even worse SM utilization  because of its data layout. The SqueezeNet variants and GoogLeNet have negligible percentage of these two kernels (Table \ref{tab:KernelResults}) and hence, $\frac{BP_t}{FP_t}$ ratio is lower than  one and remains unaffected with increasing $B$. However, MobileNet variants have both   {\tt Gemv2T} and {\tt Gemv2N}  kernel percentage substantially high (Table \ref{tab:KernelResults}) and this leads to low SM utilization in both forward and backward pass. Therefore, $\frac{BP_t}{FP_t}$ ratio is greater than one for all $B$ in MobileNet variants. Note that during backward pass, the SM utilization properties of {\tt Gemv2T} and {\tt Gemv2N} kernels get inter-changed (explained in Appendix \ref{sec:AppendixMinibatch}). That is, DNNs  with higher percentage of  {\tt Gemv2T} and lower percentage of  {\tt Gemv2N}  will have better SM utilization in forward pass, but poor SM utilization in backward pass. For the same reason, $\frac{BP_t}{FP_t}$ ratio increases with $B$ in  all the variants of SqueezeNext (Figure \ref{fig:MemfpAndFPbpRatio}(b) and \ref{fig:MemfpAndFPbpRatio}(d)). 

\begin{enumerate}  []
\item [] {\bf Observation 3:}\label{obs:timingSCA2} The increase in $T_p$ with higher $B$ depends on the SM utilization in forward pass. Both MobileNet variants and ShuffleNet variants have significantly higher percentage of {\tt Gemv2T} and {\tt Gemv2N} kernels (Table \ref{tab:KernelResults}) and hence, they have poor SM utilization in forward pass. For the same reason, $T_p$ remains constant for both MobileNet variants (for all $B$) and ShuffleNet variants (for higher $B$). 
\end{enumerate}

\begin{table} [htbp]
\caption{Characteristics of compact DNNs. $P_{norm}$ shows the reduction in number of  parameters of a DNN compared to AlexNet. Top-1 is the single crop image classification accuracy on ILSVRC 2012 dataset. }
\label{tab:Modelattributes}
\centering
\resizebox{1.0\textwidth}{!}{
\begin{tabular}{ |c| c| c| c| c| c| c| c | c| c| } 
 \hline
 \textbf{Model Name} & \textbf{Image size} & \textbf{ $M_c$} & \textbf{ $P$}  &  \textbf{$A$}  & \textbf{$A$/$P$} & \textbf{$M_c$/$P$} & \textbf{$M_c$/$A$} &  {\bf Top-1 accuracy (\%)} &  $P_{norm}$\\
 \hline

 AlexNet \cite{NIPS2012_4824}            & $224\times 224$ &  723  & 60.97  &  2.05 & 0.03  &11.86  & 352.65  &57.20  &$1\times$  \\
 \hline
 SqueezeNet-V1.0     & $224\times 224$ & 848   & 1.25   & 12.3 & 9.84 &678.08 & 68.91 &57.50  &$49\times$   \\
 SqueezeNet-V1.1   & $224\times 224$ & 349   & 1.24   &  7.2 & 5.81  &281.57 & 48.49 &57.10  &$49\times$       \\
 \hline
 1.0-G-SqNxt-23     & $224\times 224$ & 221    & 0.54   & 17.81  & 32.80 &406.35 & 12.39  &57.16      &$112\times$   \\
 1.0-SqNxt-23       & $224\times 224$ & 273    & 0.72   & 17.81  & 24.84 &380.50 & 15.32 &59.05      &$85\times$    \\
 1.0-SqNxt-23v5     & $224\times 224$ & 225    & 0.93   & 14.06 & 15.12 &242.04 & 16.01 &59.24      &$66\times$       \\
 2.0-SqNxt-23        & $224\times 224$ & 726    & 2.36   & 32.21 & 13.65 &307.62 & 22.54  &67.18      &$26\times$     \\
 2.0-SqNxt-23v5    & $224\times 224$ & 703    & 3.22   & 24.66 & 7.66  &218.41 & 28.52  &67.44      &$19\times$     \\
 \hline
 MobileNet-V1       & $224\times 224$ &  574   & 4.23   & 20.32 & 4.80  &135.65 & 28.24  &70.60      &$14\times$     \\
 MobileNet-V2       & $224\times 224$ &  300   & 3.40   & 35.45 & 10.43  &88.24 & 8.46       &72.00    &$18\times$\\
 \hline
 DenseNet-121     & $224\times 224$ &  3080  & 7.98   & 69.99 & 8.77  &385.96 & 44.01        &75.00   &$8\times$\\
 \hline
 GoogLeNet           & $224\times 224$ & 1590   & 7.00      & 10.06 & 1.44  &227.14 & 158.05  &71.00    &$9\times$    \\
 Inception-V2       & $231\times 231$ & 2200   & 11.19  & 18.03 & 1.61  &196.60 & 122.02   &76.60          &$5\times$    \\
 \hline
\end{tabular} }
\end{table}

\subsection{Power Side-channel Attack} 
This side-channel attack is based on the understanding of $EPF$ and its variation with $B$. As shown in Equation \ref{eqn:EnergyEfficinecy}, $EPF$ depends on  both $M_c$ and energy efficiency of MAC operation ($E_e$).  The former depends on DNN's architecture and design methodologies, while the latter depends on the degree of data reuse exploitable in the network. Further, the variation of $EPF$ with  $B$ also depends on resource utilization because increasing $B$ increases data-level parallelism. As explained in Appendix \ref{sec:AppendixMinibatch}, $E_e$ has strong dependence on activation reuse ($\frac{M_c}{A}$) which is significantly low in SqueezeNext variants and MobileNet variants (Table \ref{tab:Modelattributes}). The reason for low activation reuse in the former is the presence of all the fine-grained architectural components which increases the concurrent activations and reduces the effective $\frac{M_c}{A}$. The lower activation reuse in MobileNet variants is due to the use of DWConv and PWConv.

\begin{enumerate} 
\item [] {\bf Observation 4:} \label{obs:EPF_WithB} At $B$ = 1, the $EPF$ of SqueezeNext variants and MobileNet variants are very high (Figure \ref{fig:ThroughputEnergy}(b) and \ref{fig:ThroughputEnergy}(d)) due to the aforementioned low $\frac{M_c}{A}$. However, because of the poor SM utilization in the MobileNet variants, $EPF$  remains constant with increasing $B$ (Table \ref{tab:KernelResults}). By contrast, in SqueezeNext variants, due to better SM utilization, $EPF$  decreases with increasing $B$.
\end{enumerate}

\subsection{Kernels Side-channel Attack}  
\begin{figure}[htbp]
\centering
\includegraphics[scale=0.35]{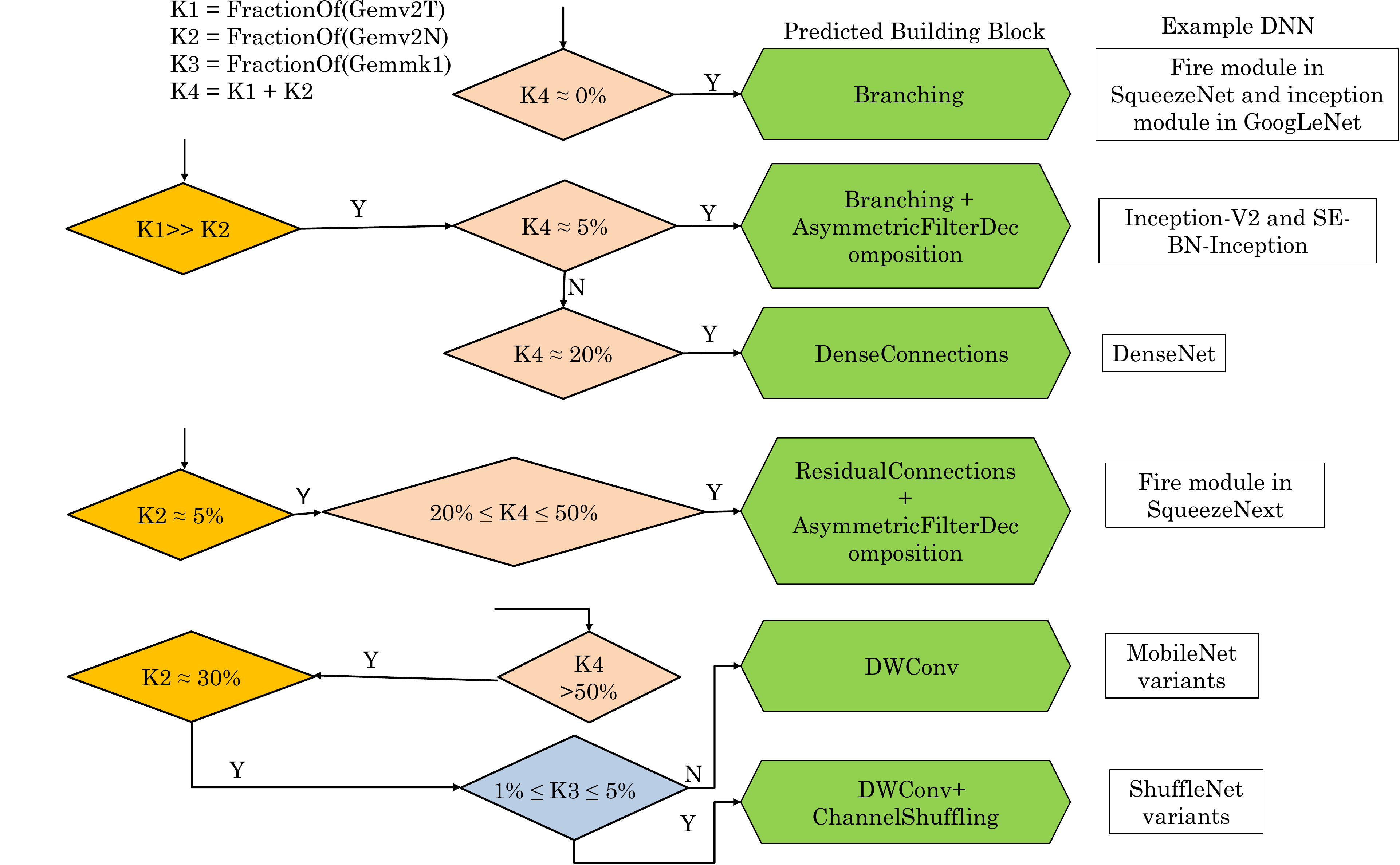}  

\caption{Prediction steps for determining fine-grained architectural ingredients of principle building blocks}
\label{algo:FineGrain}
\end{figure}

Figure \ref{algo:FineGrain} shows how an adversary can decipher the fine-grained architectural components in building blocks of compact DNNs by knowing the percentage of cuBLAS kernels {\tt Gemmk1}, {\tt Gemv2T}, and {\tt Gemv2N}.

When the basic building block has only branching, such as the fire module in SqueezeNet and inception module in GoogLeNet, then the combined contribution of {\tt Gemv2T} and  {\tt Gemv2N} is nearly zero (Table \ref{tab:KernelResults}). In comparison, the presence of asymmetric filter decomposition and branching, for example, in Inception-V2 and SE-BN-Inception, cause irregular data access and computation patterns. This results in lower SM utilization, which is substantiated by the presence of {\tt Gemv2T} ($\simeq$ 5\%) kernel (Table \ref{tab:KernelResults}). 
Further, dense connections (in DenseNet) result in higher percentage contribution of {\tt Gemv2T} ($\simeq$ 20\%). When the basic building block contains asymmetric filter decomposition along with residual connections (fire module in SqueezeNext), then it leads to a very high percentage of {\tt Gemv2T} compared to {\tt Gemv2N}, but both are significant. The presence of {\tt Gemv2T} and {\tt Gemv2N} are significantly higher when  DWConv (MobileNet variants) employed in DNNs, which renders most of the SM under-utilized. 
Note that the cuBLAS library is an integral part of CUDA \cite{cuBLAS_libraries}, which is used in most of the DNN frameworks such as Caffe, PyTorch, TensorFlow, and others. Hence, the above architectural analysis on the correlation between the fine-grained architectural components and the percentage of cuBLAS kernels is quite generalized and distinctive. Thus, it can be used by an adversary to predict the primary building block in DNNs.

{\em We utilize the observations made above in both inter-group prediction and intra-group prediction stages of the ``DeepPeep'' attack model (Section \ref{subsec:InterGropPrediction} and Section \ref{subsec:IntraGropPrediction}) to decipher the architecture of compact DNNs.}

\section{Proposed Attack Model: DeepPeep} \label{sec:AttackModel}

\begin{figure} [htbp] \centering
\includegraphics[scale=0.36]{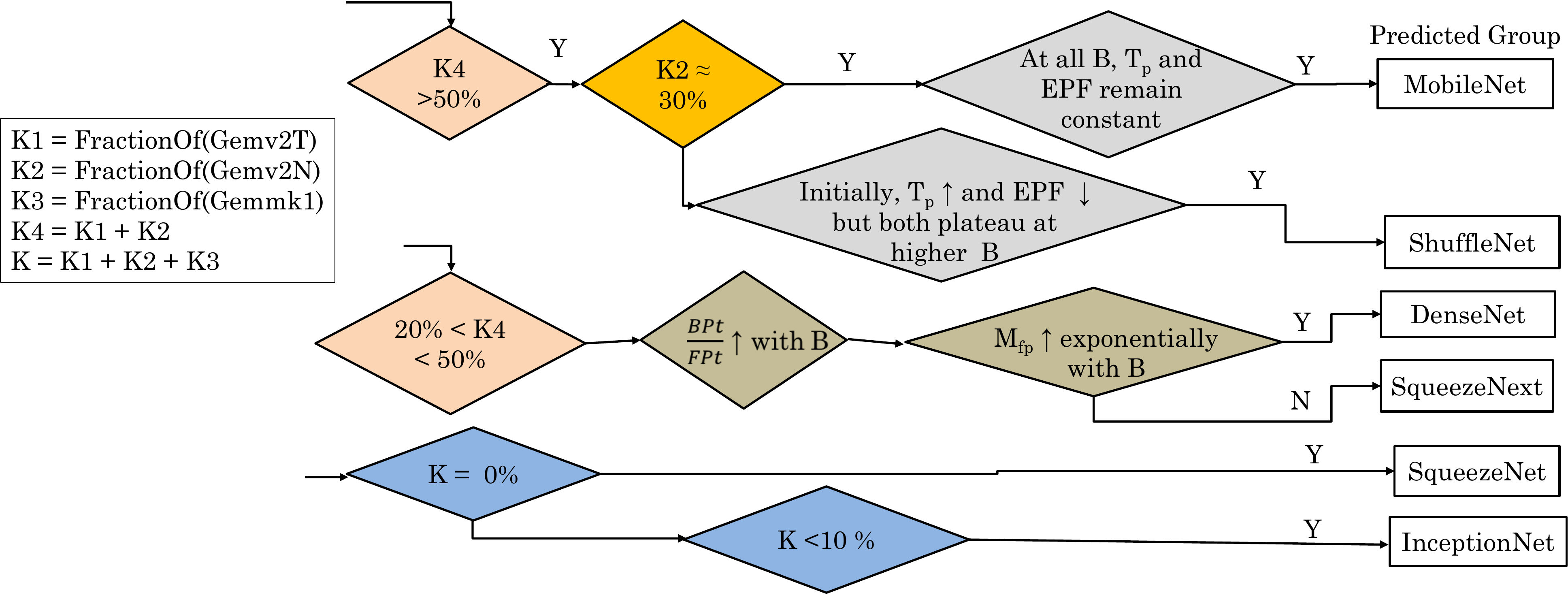}
\caption{Inter-group prediction stage in DeepPeep predicts the group of victim DNNs} 
\label{fig:InterGroupDeepPeep}
\end{figure}

In this section, we describe our proposed two-stage attack methodology, ``DeepPeep,'' which exposes the crucial security vulnerabilities present in compact DNNs due to its design principles. ``DeepPeep,''  decipher the architecture of the compact DNN (termed ``victim DNN'') running on GPU in a cloud. Based on the similarity of the building block, we group the compact DNNs into six groups, as shown in  Table \ref{tab:CDNNblocks}. In each group, we take one DNN as the reference, which is highlighted as bold in Table \ref{tab:CDNNblocks}. The attack proceeds in two stages: 
\begin{enumerate}
\item Inter-group prediction:  First,  the adversary finds  the group of the victim DNN (Figure \ref{fig:InterGroupDeepPeep}) 
\item Intra-group prediction: Once the group is known, the adversary compares the characteristics, such as $FP_t$, $BP_t$, $M_{fp}$, $EPF$ (at different $B$), percentage of cuBLAS kernels (Figure \ref{algo:FineGrain}), etc., of the reference DNN with the victim DNN (Figure \ref{fig:IntraGroupDeepPeep}).  Note that the inter-group prediction is easier than intra-group prediction because the characteristics of building blocks vary significantly across the DNNs, while the intra-group variation is modest.
\end{enumerate}

\subsection{Inter-group Prediction in DeepPeep} \label{subsec:InterGropPrediction}
Figure \ref{fig:InterGroupDeepPeep} summarizes the steps for inter-group prediction stage of DeepPeep attack model. In this prediction stage, the inputs are $T_p$, $EPF$, $BP_t$, $FP_t$, $M_fp$, and percentage contribution of cuBLAS kernels {\tt Gemv2T}, {\tt Gemv2N} and {\tt Gemmk1} (Table \ref{tab:KernelResults}). The outputs of inter-group prediction stage are the DNN groups listed in Table \ref{tab:CDNNblocks}. We show the $T_p$ and $EPF$ values in Figure \ref{fig:ThroughputEnergy}, and   $M_{fp}$ and $\frac{BP_t}{FP_t}$ ratio in Figure \ref{fig:MemfpAndFPbpRatio}.  We experimentally demonstrate the variation of aforementioned metrics with the batch size ($B$) on both P100 and P4000 GPUs.

\textbf{Case 1:} As described in Figure \ref{fig:InterGroupDeepPeep},  when the combined contribution of {\tt Gemv2T} and  {\tt Gemv2N} kernels is very high,  then it confirms the presence of DWConv with PWConv. Hence, the victim DNN belongs to either the MobileNet group or the ShuffleNet group. In the MobileNet group, the majority of operations are DWConv and PWConv, which lead to poor utilization of SM, and hence,  $T_p$ and $EPF$ remain constant even with higher $B$ (Figure \ref{fig:ThroughputEnergy}). By comparison, in the ShuffleNet group, each layer has channel shuffling operation, which leads to better SM utilization compared to the MobileNet group and the total percentage of {\tt Gemv2T} and  {\tt Gemv2N} kernels is lower than DNNs in MobileNet group (Table \ref{tab:KernelResults}). Also, the percentage contribution of {\tt Gemmk1} kernel in DNNs of the ShuffleNet group is significantly higher than that in DNNs of the MobileNet group. Due to this, for ShuffleNet DNNs, with increasing $B$, initially $T_p$ increases, and $EPF$ decreases, but they plateau at higher $B$.

\textbf{Case 2:} If total contribution of {\tt Gemv2T} and {\tt Gemv2N} kernels is lower than that in Case 1 and the ratio $\frac{BP_t}{FP_t}$ increases at higher $B$, then the victim DNN belongs either to SqueezeNext group or DenseNet group. This has been explained in Appendix \ref{sec:AppendixMinibatch} and is  illustrated in Figures \ref{fig:MemfpAndFPbpRatio}(b) and \ref{fig:MemfpAndFPbpRatio}(d). Further, if $M_{fp}$ increases exponentially with $B$, it confirms the presence of dense connections in principle building block, and hence, the victim DNN belongs to the DenseNet group. By contrast, if  $M_{fp}$ shows an only linear increase with increasing $B$, then it indicates that there are fewer skip/residual connections, and the victim DNN belongs to the SqueezeNext group.  

\textbf{Case 3:} When the combined contribution of all the three cuBLAS kernel {\tt Gemv2T}, {\tt Gemv2N} and {\tt Gemmk1} is lower than 10\%, then victim DNN belongs to either SqueezeNet or InceptionNet group. If the above-mentioned combined contribution is less than 1\%, then the victim DNN is either GoogLeNet or SqueezeNet. 

By following the above steps, an adversary can easily predict the group of a DNN, especially when the characteristics of the groups are very distinct, for example, MobileNet and DenseNet.  However, the characteristics of some groups such as  SqueezeNet and InceptionNet, are quite similar, and hence, more investigation is required for accurately distinguishing between them.

\subsection{Intra-group Prediction} \label{subsec:IntraGropPrediction}
\begin{figure} [htbp] \centering
\includegraphics[scale=0.40]{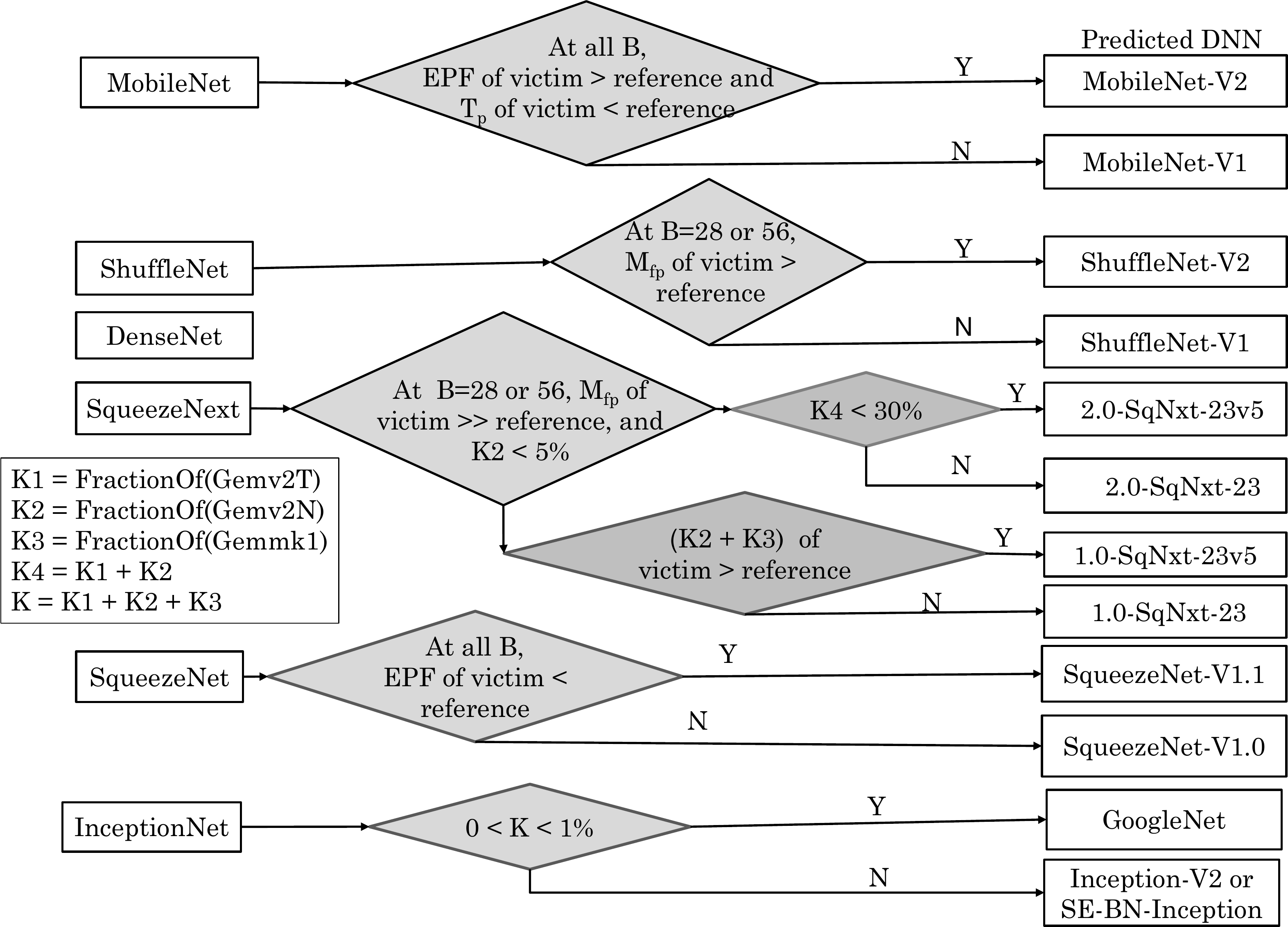}  
\caption{Intra-group prediction stage in DeepPeep predicts the exact DNN in the group} \label{fig:IntraGroupDeepPeep}  
\end{figure}

Once the group of a DNN is known, we further predict the exact DNN in the group, through side-channel attacks. We choose one DNN in each group as the reference DNN (shown in bold in Table \ref{tab:CDNNblocks}). In the intra-group prediction stage of DeepPeep attack model, shown in Figure \ref{fig:IntraGroupDeepPeep}, we compare the characteristics of the victim DNN with those of the reference DNN. Since DNNs in most of the groups exhibit very distinct characteristics at a larger $B$, they are easily distinguishable through side-channel attacks. Since the DenseNet group has only one DNN, there is no need to perform intra-group prediction. 

\begin{itemize}
\item \textbf{MobileNet group: } If $EPF$  and $T_p$ of victim DNN is higher and lower (respectively) than that of the reference DNN at all values of $B$, then victim DNN is MobileNet-V2 (Figures \ref{fig:ThroughputEnergy}). Apart from thses distinguishing performance metrics, $M_{fp}$ is also a distinguishable factor here because  MobileNet-V2 has substantially higher $M_{fp}$ compared to MobileNet-V1 at higher $B$.

{\em Reasoning}: 
In the primary building block of MobileNet-V2, the DWConv layer is sandwiched between two PWConv layers and has one residual connection \cite{8578572}. These residual connections increase concurrent activations. MobileNet-V2 is quite deeper than MobileNet-V1, which further increases the total concurrent activations and results in very high $M_{fp}$  in MobileNet-V2. Also, this reduces effective activation reuse and increases energy consumption.

\item \textbf{ShuffleNet group: }  
If $M_{fp}$ of victim DNN is higher than that of the reference DNN at higher $B$, then it is ShuffleNet-V2.

{\em Reasoning}: The only architectural difference between the building blocks of ShuffleNet-V1 and ShuffleNet-V2 is that unlike the former, the latter employed 1) group convolution in  $1\times1$ convolutional layers and 2)  channel shuffling after the DWConv \cite{Ma_2018_ECCV}. This architectural difference leads to a difference between their $M_{fp}$, which is more pronounced at higher $B$. 

\item \textbf{SqueezeNext group: } If $M_{fp}$ of victim DNN is very high compared to reference DNN at higher $B$ (such as $B$=28 or 56), (Figures \ref{fig:MemfpAndFPbpRatio}(a) and  \ref{fig:MemfpAndFPbpRatio}(c)) and also the  percentage of {\tt Gemv2N} kernel is lower than 5\% then victim DNN is either 2.0-SqNxt-23 or 2.0-SqNxt-23v5. Otherwise, it could be either 1.0-SqNxt-23v5 or 1.0-G-SqNxt-23. Further, if the combined contribution of {\tt Gemv2T} and {\tt Gemv2N} kernels is lower than 30\%, then victim DNN is 2.0-SqNxt-23v5, else it is 2.0-SqNxt-23 (Table \ref{tab:KernelResults}). To distinguish between 1.0-SqNxt-23v5 and 1.0-G-SqNxt-23, we use kernel side-channel attack   which is used to predict the fine-grained architectural components (Figure \ref{algo:FineGrain}). If the percentage contribution of both {\tt Gemv2N} and {\tt Gemmk1} kernel is higher than that of the reference DNN, then it is 1.0-SqNxt-23v5, else, it is 1.0-G-SqNxt-23 (Table \ref{tab:KernelResults}).

{\em Reasoning}: The v5 versions, 1.0-SqNxt-23v5 and 2.0-SqNxt-23v5, of SqueezeNext have a lower number of groups with high ifmap resolution in the initial layers of networks and a higher number of groups with low ifmap resolution in deeper layers \cite{Gholami2018SqueezeNextHN}. Therefore, SM utilization of v5 versions is better than their respective baseline models. Also, the percentage of {\tt Gemv2T} kernel is lower, while that of the {\tt Gemmk1} kernel is higher than their respective baseline. Further, 2.0-SqNxt-23 and 2.0-SqNxt-23v5 have $2\times$ number of filters per layer compared to 1.0-SqNxt-23 and 1.0-SqNxt-23v5 respectively, and hence, their  $M_{fp}$ is larger and easily distinguishable at higher $B$.

\item \textbf{SqueezeNet group:}  There are two distinguishing factors, $EPF$ and $M_{fp}$ for the DNNs belonging to this group. If $EPF$ of victim DNN is lower than that of the reference DNN in this group at all $B$ then the victim DNN is SqueezeNet V1.1 (Figures \ref{fig:ThroughputEnergy}(b) and  \ref{fig:ThroughputEnergy}(d)). Also, if $M_{fp}$ of victim DNN is lower than that of the reference DNN at all $B$ then it confirms the presence of SqueezeNet V1.1 (Figures \ref{fig:MemfpAndFPbpRatio}(a) and  \ref{fig:MemfpAndFPbpRatio}(c)).  These distinguishing factors become more evident at higher $B$ (such as $B$=28, 56). 

{\em Reasoning}: The architectures of SqueezeNetV1.0 and SqueezeNetV1.1 are quite similar except that V1.1 has a low dimensional filter in the first convolutional  layer and uses down-sampling earlier in layers. Downsampling in earlier layers reduces the intermediate feature map size, which reduces both $M_c$ and $M_{fp}$ and makes the network more energy-efficient.

\item \textbf{InceptionNet group: } 
If the percentage contribution of {\tt Gemv2T} kernel in victim DNN is more than 1\%, then it can be either Inception-V2 and SE-BN-Inception. However, since the architectural difference between inception-V2 and SE-BN-Inception is negligible, side-channel attacks are unable to further distinguish these two DNNs. 

{\em Reasoning}:
Except for GoogLeNet, other DNNs in this group have asymmetric filter decomposition (${AsymmFilterDecomp}$), which leads to a higher percentage of {\tt Gemv2T} kernels. Thus, GoogLeNet is easily distinguishable in this group. However, SE-BN-Inception uses very few residual connections to employ feature recalibration, which increases the representational power of the network. The small number of residual connections in SE-BN-Inception leads to a small increase in the percentage contribution of {\tt Gemv2T} kernel (Table \ref{tab:KernelResults}). Therefore, performance metrics of Inception-V2 and SE-BN-Inception are quite similar (Figures \ref{fig:MemfpAndFPbpRatio} and \ref{fig:ThroughputEnergy}) and hence, our side-channel attacks are unable to distinguish inception-V2 and SE-BN-Inception. 
\end{itemize}

\section{Defense Mechanism} \label{sec:DefenceMechanism}

In sections \ref{subsec:InterGropPrediction} and section \ref{subsec:IntraGropPrediction}, we have seen how the distinctive characteristics of architectural building blocks can be exploited to decipher the design methodologies of compact DNNs. We observed that some architectures, such as the Dense block and DWConv, could be quite easily predicted and hence have a high risk of IP theft. In this section, we propose some design guidelines which circumvent the potential risk of design IP theft. We also study the parameter and computation overheads of our secure design and demonstrate the implications for predictive performance (top-1 accuracy). Section \ref{sec:DNNDWConv} presents specific guidelines for DNNs that use DWConv, and Section \ref{sec:genericDNN} provides generic recommendations and guidelines for the security-aware design of DNNs.

\subsection{Guidelines for DNNs with DWConv}\label{sec:DNNDWConv}  
Instead of DWConv, we employ group convolution in MobileNet-V1. Conventionally, in group convolution, $g$ remains constant in all the convolutional layers \cite{Xie2017AggregatedRT,Zhang2018ShuffleNetAE}. By contrast, we employ group convolution where the number of filter channels ($g$ = $\frac{M}{G}$) remains constant in each group of a convolutional layer. Note that $M_c$ and data-reuse depend on $g$ \cite{jha2020e2gc}, and having the same $g$ in all the convolutional layers of DNNs can result in similar characteristics which can be exploited by malicious entities to extract the design information. Moreover, compared to group convolution with constant $G$, group convolution with constant $g$ incurs significantly higher computational and parameter overheads \cite{jha2020e2gc}. Hence, we employ group convolution with a constant $G$ in MobileNet-V1, where the value of $g$ depends on the number of ifmaps (Table \ref{tab:SecureMobileNet}).  We vary $G$ from 1 to 32 and measure the performance of each architecture in terms of $FP_t$, $BP_t$ and $M_{fp}$. Table \ref{tab:SecureMobileNetOverheads} shows the results. As shown in Table \ref{tab:SecureMobileNetOverheads}, increasing value of $G$ increases the number of parameters and computations;however, both $FP_t$ and  $BP_t$ decrease with increasing $G$.

\begin{table}[htbp]
    
        \centering
        \caption{Layers with depthwise convolution (DWConv) in MobileNet-V1}
        \label{tab:SecureMobileNet}
        \begin{tabular}[htbp]{ C{5cm} | c | c }
            \toprule
           {\bf Convolutional layers } & {\bf Convolution types }& { \bf (ifmap, ofmap, \# groups)} \\
            \toprule
            Conv2\_1 & DWConv & (32, 32, $\frac{32}{G}$)\\ \midrule
            
            Conv2\_2 & DWConv& (64,  64, $\frac{64}{G}$)\\ \midrule
            Conv3\_1 & DWConv & (128, 128, $\frac{128}{G}$)\\ \midrule
            Conv3\_2 & DWConv& (128, 128, $\frac{128}{G}$)\\ \midrule
            Conv4\_1 & DWConv & (256, 256, $\frac{256}{G}$)\\ \midrule
            Conv4\_2 & DWConv& (256, 256, $\frac{256}{G}$)\\  \midrule
            
            Conv5\_1, Conv5\_2, Conv5\_3, Conv5\_4, Conv5\_5, Conv5\_6  & DWConv & $6\times (512, 512, \frac{512}{G})$\\ \midrule
            
            Conv6 & DWConv &(1024, 1024, $\frac{1024}{G}$) \\
            
            \bottomrule
            
        \end{tabular}
 \end{table}

\begin{figure}[htbp]
\centering
\includegraphics[scale=0.50]{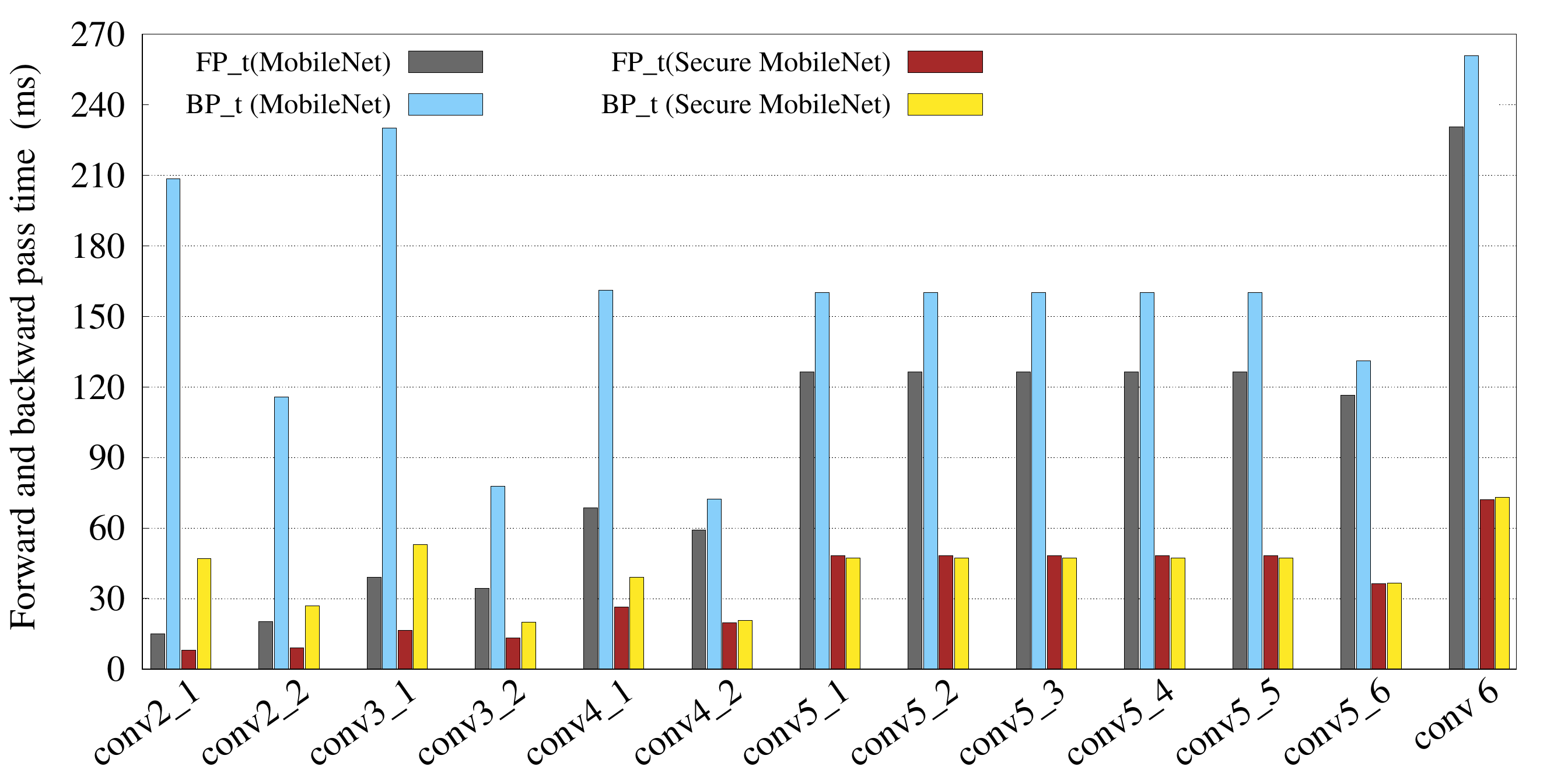} 
\caption{Variation in $FP_t$ and $BP_t$ time (measured on P100 GPU with $B$ = 56) of all the layers with DWConv in baseline MobileNet-V1 and secure MobileNet-V1 with $G$ = 4}
\label{fig:SecureMobileNet}
\end{figure}

One of the distinctive properties of MobileNets, which makes it quite predictable in timing side-channel attack, is the substantial difference between the $FP_t$ and $BP_t$ even at higher $B$ (Figure \ref{fig:SecureMobileNet}). However, increasing $G$ in MobileNet-V1 reduces the disparity between $FP_t$ and $BP_t$ (Table \ref{tab:SecureMobileNetOverheads}). This happens because having a larger $G$ reduces the fragmented memory accesses and increases the data reuse, which reduces the latency of MAC operations. We select an optimum $G$ as $G$=4, where both the computational and parameter overheads are optimal along with the reasonable  (layer-wise) disparity between  $FP_t$ and $BP_t$. The lower disparity between $FP_t$ and $BP_t$ in MobileNet-V1 with $G$=4 makes the non-trivial to predict the architecture of building block through timing side-channel attack. We call this version of MobileNet-V1 with $G$=4 as {\em secure MobileNet-V1}.  

\begin{table}[htbp]\centering
\caption{Computation and parameter overhead of ``secure MobileNet'' over baseline MobileNet (measured  on P100 GPU with $B$=56)}
 \label{tab:SecureMobileNetOverheads}
         \begin{tabular}[htbp]{ |c | c | c | c | c | c| }
   \hline
 {\bf $G$ } & {\bf Increase in $M_c$ (\%)}&  {\bf Increase in $P$ (\%)}& {\bf $FP_t$ (ms)  }&{\bf $BP_t$ (ms) } &{\bf $M_{fp}$ (MB)}  \\
            \hline
            
             1 & - & - & 1286 & 2143 & 7206 \\
             2 & 3.0 & 1.2 & 760 & 1297 & 7206 \\
             4 & 8.8 & 3.3 & 506 & 645 & 7206 \\
             8 & 19.4 & 7.3 & 285 & 493 & 7206 \\
             16 & 37.6 & 15.8 & 206 & 306 & 7204 \\
             32 & 64.2 & 32.9 & 174 & 213 & 7204 \\ 
            
            \hline
        \end{tabular}
 \end{table}

{\bf Performance implications of designing secure MobileNet-V1:} Secure MobileNet-V1 has only 8.8\% and 3.3\% additional $M_c$ and $P$ respectively, and compared to baseline MobileNet-V1 latency is improved by 60.6\% (Figure \ref{fig:SecureMobileNet}). Notice that changing $G$ does not change  $M_{fp}$.  Furthermore, we investigate the predictive performance implication of designing secure MobileNet-V1. We trained baseline MobileNet-V1 and MobileNet-V1 with $G=4$ on there image classification datasets ImageNet-1K \cite{Russakovsky:2015:ILS:2846547.2846559}, CIFAR-100 \cite{2012KrizhevskyCIFAR} and Imagenette \cite{Fastai_Imagenette}. Results are shown in Table \ref{tab:SecureMobileNetAccuracy}. The top-1 accuracy on two complex datasets ImageNet-1K and CIFAR-100, have increased by $\approx$2\%. However, on the easy dataset, Imagenette, the top-1 accuracy is improved by 0.8\%. Increasing $G$ increases the number of channels in a group and captures more variations of a concept; thus, it boosts the network's representational power. This phenomenon is more pronounced in complex datasets (such as ImageNet-1K and CIFAR-100), where capturing a higher number of variations improves the predictive performance significantly. As a result, compared to Imagenette, the gain in accuracy is higher on  ImageNet-1K and CIFAR-100 dataset (Table \ref{tab:SecureMobileNetAccuracy}). 
 
\begin{table}[htbp]\centering
\caption{Comparison of Top-1 accuracy (image classification) on ImageNet-1K, CIFAR-100, and Imagenette datasets}
 \label{tab:SecureMobileNetAccuracy}
         \begin{tabular}[htbp]{c | c | c | c }
   \toprule
 {\bf DNN model } & {\bf ImageNet-1K (\%)}&  {\bf CIFAR-100 (\%)}& {\bf Imagenette (\%) }  \\
            \toprule
             Baseline MobileNet-V1 ($G$=1) & 70.60 & 67.44 & 84.12  \\ \midrule
             Secure MobileNet-V1 ($G$=4) & 72.24 & 69.15 & 84.92 \\
                 
            \bottomrule
        \end{tabular}
 \end{table}

\subsection{Guidelines for any generic DNN}\label{sec:genericDNN} 
In this section, we present some other architectural heuristics, as generic recommendations, which can be applied to thwart the side-channel attacks and enable robust DNNs. These design techniques can be incorporated into a diverse set of DNNs.

\textbf{1. Using a mixture of building blocks in DNN's design: } 
We have shown that (1) dense connections in dense blocks lead to an exponential increase in $M_{fp}$ with increasing $B$, and (2)  DWConv lowers the computing resource utilization, which cannot be mitigated with higher $B$. Therefore, the blocks showing completely different behavior with increasing $B$ can be combined to thwart IP theft through side-channel attacks.  Compared to initial layers, deeper layers (layers towards output) have more higher-dimensional semantic information and less positional information \cite{li2019frd}; consequently, a higher number of filters is required in deeper layers. Therefore, a secure version of compact DNNs should use dense blocks in the deeper layers and improve the feature reuse, while DWConv in the initial layers.  These types of mixture building blocks in DNN reduce computational complexity and boost the representational power using improved feature reuse. While the absence of distinctive properties in DNNs makes it non-trivial to decipher the architecture through side-channel attacks. 

Note that employing skip connects only in deeper layers could result in diminished predictive performance. However, the lower number of skip connections leads to lower $M_{fp}$, and the network can fit into the limited on-chip memory. This further reduces the data movements and improves the energy-efficiency and throughput of the network.

\textbf{2. Inclusion of sophisticated design heuristics in existing building blocks: } To make intra-group prediction difficult, design heuristics such as channel shuffling and filter recalibration can be incorporated in the existing blocks, such as fire module. {\em These heuristics do not significantly affect the characteristics of building blocks in DNNs, but increase the representational power of the network. This makes it more difficult to  distinguish the DNNs with similar architecture.} 

Channel shuffling is employed after the group convolutional layer in DNN to blend the information from different groups. This prevents information loss and hence the predictive performance. However, channel shuffling is useful only up to a smaller value of $g$ \cite{Zhang2018ShuffleNetAE}. Therefore, increasing $g$ decreases the computational complexity and the number of parameters. However, beyond a specific value of $g$, channel shuffling will not be effective in recovering the information loss across different groups in the convolutional layers.  

In summary, DNN designers need to employ different design heuristics for maximizing feature reuse and data reuse (for better predictive performance and energy-efficiency, respectively) such that the network does not exhibits distinctive characteristics. Hence, deciphering the architectural components through side-channel attacks would become increasingly difficult for an adversary.

\section{Generality and applicability of ``DeepPeep''} \label{sec:GeneralityOfModel}

In this section, we discuss and analyze the applicability of DeepPeep on other DNN architectures (Section \ref{sec:extensionOtherDNNs}). We also describe the effect of datasets (inputs to the DNNs) on the efficacy of the DeepPeep attack model (Section \ref{sec:OtherDatasets}). Further, we discuss the generality of ``DeepPeep'' on other hardware platforms such as CPU and systolic-architecture based hardware accelerators (Section \ref{sec:otherHardware}).  

\subsection{Applicability on Other DNN Architectures}\label{sec:extensionOtherDNNs} 
Here, we seek to answer the following questions.  {\em Does our attack consider all the existing architectures in (manually designed) compact DNNs? Further, is DeepPeep applicable to the compact DNNs designed through an automated process such as neural architecture search (NAS)?}

There are two broad strategies for designing compact DNNs. The first strategy seeks to significantly reduce the computational complexity and/or the number of parameters with a negligible loss in predictive performance. Examples of this strategy include DWConv in MobileNet and ShuffleNet variants, and fire module in SqueezeNet and SqueezeNext variants. The second strategy boosts the representational power of DNNs in a fixed budget of MAC operations and the number of parameters. The representational power of DNNs is improved either by improving feature reuse or through multi-scale feature learning. DNNs employ residual connections  (element-wise addition) and skip connections (concatenation of fmaps from previous layers) to improve the feature reuse \cite{li2019frd}. Residual connections have been implemented in an aggressive version of the fire module, which forms the building blocks of SqueezeNext variants \cite{Gholami2018SqueezeNextHN}. Skip connections have been implemented in the dense blocks of DenseNet \cite{8099726}. Evidently, the compact DNNs used in our experiments (Table \ref{tab:CDNNblocks}) are employed all the commonly used design heuristics in the architecture of building blocks. Besides, we have included compact DNNs that use design techniques, such as channel shuffling, which are not commonly used in the design of compact DNNs. In summary, the compact DNNs used in our experiments (Table \ref{tab:CDNNblocks}) employ an extensive set of design heuristics, and hence, they are a good representative of state-of-the-art compact DNNs. 

In the automated design process such as NAS, normal and reduction cells are repeatedly stacked (alternatively) in a steam-lined fashion \cite{Zoph_2018_CVPR}. An optimization algorithm selects the architectural components of these cells over a large search space of design heuristics. This search space includes: convolution, depthwise-separable convolution, max/avg-pooling, asymmetrical filter decomposition with different filter size, identity connections such as residual/skip connections, etc. \cite{Zoph_2018_CVPR}. Since our attack model ``DeepPeep'' is a bottom-up approach that first predicts the fine-grained architectural components and then predicts the building blocks in DNNs, it can predict the normal/reduction cell in the DNN designed through NAS. In summary, the attack methodologies in ``DeepPeep'' are quite generalized and applicable to any DNN formed by repeated stacking of a building block. Hence, DeepPeep can also work on compact DNNs designed using NAS.  

\subsection{Applicability on different Datasets}\label{sec:OtherDatasets}
In the adversarial attacks \cite{goodfellow2014explaining}, input pixels are intelligently manipulated in a manner that is imperceptible to the human, to change the output prediction. By contrast, ``DeepPeep'' is entirely based on the side-channel attacks. Consequently, the input pixel values or the input dataset's complexity do not affect any intermediate steps in inter-group and intra-group prediction. Changing the spatial size of input frame, for instance $224\times224$ in ImageNet-1K vs. $32\times32$ in CIFAR-100, would change the absolute values of $M_{fp}$, $FP_t$, $BP_t$, $T_p$, and $EPF$. However, the relative trend of these performance metrics with different $B$ would remain the same. Thus, we do not need to adapt the intermediate steps in inter-group and intra-group algorithms, and hence ``DeepPeep'' attack model is independent of the input dataset.

\subsection{Applicability on Other Hardware Platforms}\label{sec:otherHardware}

We have conducted experiments on two GPUs, P4000 and P100, to show whether the distinctive characteristics of various architectural components in DNNs are consistent over GPUs with different configurations.  As shown in Figure \ref{fig:MemfpAndFPbpRatio} and Figure \ref{fig:ThroughputEnergy}, the absolute values of $M_{fp}$, $FP_t$, $BP_t$, $T_p$, and $EPF$ are different on different GPUs. However, the trends in their variation with different $B$ are consistent across these GPUs. The intermediate steps in both inter-group and intra-group prediction depends on the aforementioned trends with different $B$ (Figure \ref{fig:InterGroupDeepPeep} and \ref{fig:IntraGroupDeepPeep}). Moreover, the prediction of fine-grained architectural ingredients of building blocks (Figure \ref{algo:FineGrain}), which serves as an intermediate step in both inter-group and intra-group prediction, depends entirely on the percentage of cuBLAS kernels. Hence, our proposed attack method is GPU-agnostic. 

On CPU, DNN inference heavily relies on the tiled GEMM kernel, and an adversary can exploit this relationship to gain insights into the DNN architecture \cite{2020_USENIX_Yan}. On GPUs, the hardware utilization is observed based on the percentages of three cuBLAS kernels {\tt Gemv2T}, {\tt Gemv2N}, and {\tt Gemmk1}. On CPUs, a different set of kernels need to be observed for finding the hardware utilization. These kernels depend on the library used, e.g., Intel MKL-DNN. Hence, to use the ``DeepPeep'' attack model on CPU, we need to adapt the intermediate steps in both inter-group and intra-group prediction.  Similarly, on systolic-array based DNN accelerators, the percentage of PE (processing element) utilizations can be used as an intermediate step in the ``DeepPeep'' attack algorithm. However, this information may be insufficient to decipher all the fine-grained architectural ingredients in principle building block. We believe that we need to change the intermediate steps in both inter-group and intra-group algorithms to use ``DeepPeep'' on the systolic accelerators.

\section{Reasons for Human-Observation Based Approach in ``DeepPeep''} \label{sec:Discussion}
 
Unlike the machine learning-based approach, ``DeepPeep" is a human observation based attack model where the insights obtained from the correlation between architectural components and their performance implications have been exploited to reverse engineer the architecture of compact DNNs. Therefore, the steps used in deciphering fine-grained architectural components (Figure \ref{algo:FineGrain}) have been used as intermediate steps in both stages Inter-group and Intra-group prediction in ``DeepPeep". The reasons for preferring a human-observation based approach over a machine learning-based approach is threefold.

{\bf 1. Problem structure:} Unlike a simple composite function or a sequence-based function, the ``DeepPeep" algorithm is a {\em relational reasoning} problem where the intermediate steps  in Figure \ref{algo:FineGrain}, Figure \ref{fig:InterGroupDeepPeep}, and Figure \ref{fig:IntraGroupDeepPeep} resembles nodes in a tree/graph. Hence, a simple feed-forward neural network (MLP or CNN) or RNN would not generalize well over such a problem. For (sample efficient) generalization, the computational graph of the network should very-well align with the structure of the relational reasoning problem \cite{Xu2020What}. Therefore, a more structured graph neural network (GNN) would be an appropriate algorithm. However, the current understanding of the relationship between generalization ability and structure for reasoning is quite limited \cite{Xu2020What}. \\
{\bf 2. Sample size:} Neural networks require a huge dataset to learn and develop the experience. Transfer learning can mitigate this issue; however, it is limited to only correlated problems. That is, neural networks are quite inefficient, as compared to a human, to find the relationship between entirely different experiences. As shown in Table \ref{tab:CDNNblocks}, we have very few data points (DNNs) for each fine-grained architectural components and building blocks. \\   
{\bf 3. Lack of reasoning:} One of the crucial shortcomings of the neural network is a lack of reasoning. Note that in the ``DeepPeep" attack, we reverse engineer the architecture of compact DNNs using the information obtained from side-channel attacks. Hence, a human-based approach is preferred for such tasks with very-limited data points.

\section{Related Work} \label{sec:RelatedWork} 

\textbf{Related work on deciphering DNN architecture:} 
Liu et al. \cite{Liu:2018:SAE:3201607.3201772}, investigate the security challenges in model compression. Liu et al. \cite{8383920} expose security vulnerabilities through the Trojan attacks. Hua et al. \cite{8465773} exploit the memory access pattern to decipher the architecture of AlexNet and SqueezeNet on a hardware accelerator.  Yan et al. \cite{2020_USENIX_Yan} use {\tt Gemm} library calls and dimensions of matrices in {\tt Gemm} for extracting the model information. They performed experiments on VGG and ResNet. However, these studies perform experiments on a limited set of DNNs with simple architectures; they have not shown the effectiveness of their technique on complex design methodologies, such as DWConv, channel shuffling, dense connections. Also, some of these methods require privileged access to the hardware/software. Unlike the aforementioned works, we investigate the security implication of designing compact DNNs and show the feasibility of design IP theft. In our experimental evaluation, we include DNNs with complex architecture such as MobileNet variants (DWConv), DenseNet (dense connections), ShuffleNet variants (DWConv and channel shuffling). Moreover, our attack methodology ``DeepPeep" uses side-channel attacks along with cuBLAS kernel analysis and hence, does not require privileged access.

\textbf{Related work on IP protection:} Rouhani et al. \cite{2019_ASPLOS_DeepSigns} proposed a technique for enabling end-to-end IP protection of DNNs using a digital watermark. Tramer et al. \cite{tramer2018slalom} proposed a  trusted execution environment for DNN inference to ensure integrity and privacy. However, they did not address the model's privacy, and the effectiveness of their technique during the training has not been proven. Also, these works are orthogonal to our work and can be integrated with our techniques.

\section{Conclusion and Future Work} \label{sec:conclusion}
In this paper, we study the ramifications of design heuristics employed in state-of-the-art compact DNNs and show how these ramifications can be exploited by malicious entities to decipher the design methods incorporated in DNN and steal the design IP. We show how some design methods, such as dense connections and DWConv, are quite easy to decipher because of their unique characteristics. We propose a novel and lightweight design guidelines for preventing information leakage in secure DNNs.  

{\bf Future work:}
We plan to modify the ``DeepPeep'' such that we can also predict the architecture of a DNN, which does not have a regular stack of similar building blocks. Also, the networks generated through random network generator \cite{Xie_2019_ICCV} have irregularity in terms of skip connections. Predicting the architecture of such an irregular network is quite challenging. In addition, we plan to incorporate the intermediate steps such that the number of convolutional layers, shape, and size of filters in convolutional layers can also be accurately predicted. Furthermore, the information about processing-element (PE) utilization and the number of systolic array calls during the execution of a DNN on the systolic array-based processor can be incorporated in the inter-group and intra-group prediction algorithms. This will make ``DeepPeep''  efficacious on all the systolic array-based DNN accelerators.

{
\bibliographystyle{ACM-Reference-Format}
\bibliography{Ref}
}

\newpage
\appendix
\renewcommand{\thesection}{\Roman{section}}

\section{Minibatch Stochastic Gradient Descent} \label{sec:AppendixMinibatch}

Here we explain the reason for the substantial difference between $FP_t$ and $BP_t$ when DWConv is employed in the design of compact DNNs (Figure \ref{fig:SecureMobileNet}). We also explain why with higher $B$, the ratio $\frac{BP_t}{FP_t}$ increases in SqueezeNext but remains constant in MobileNet and ShuffleNet. 

\begin{figure}[htbp] \centering
\includegraphics[scale=0.5]{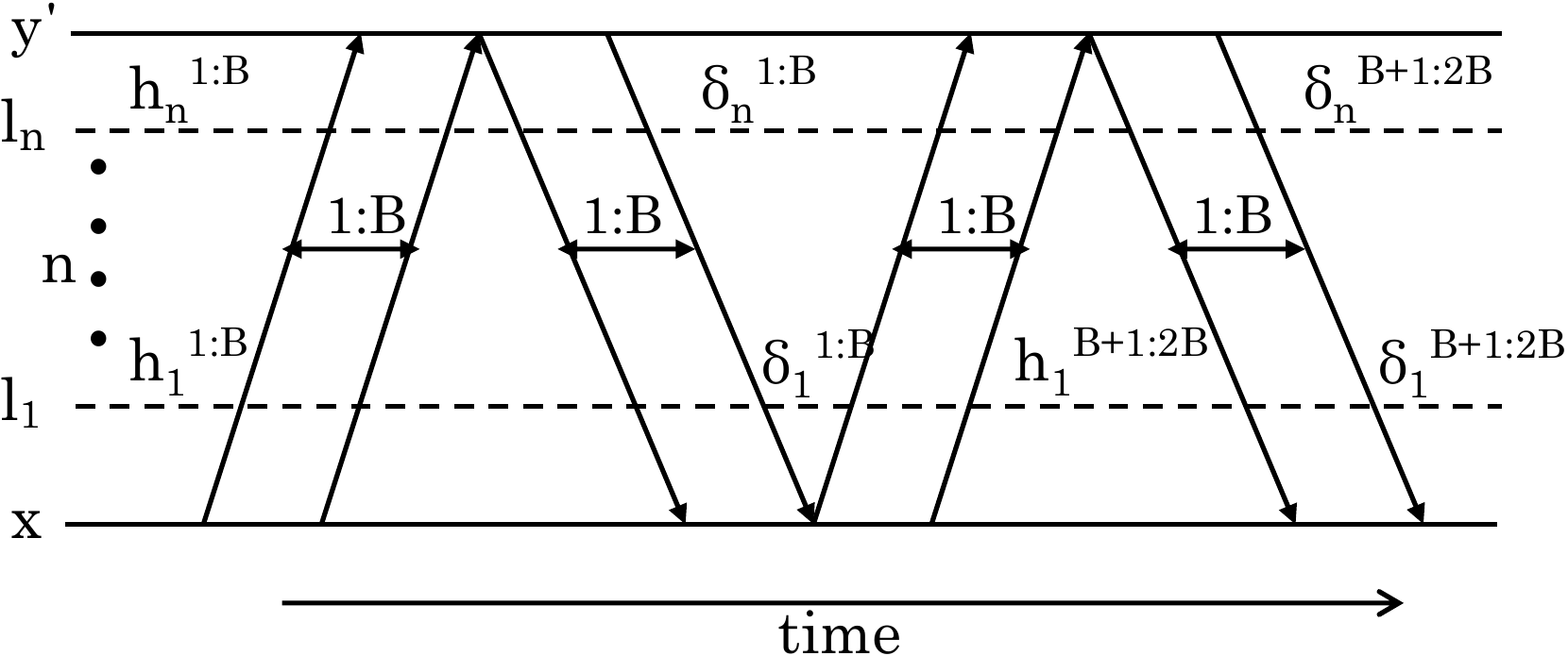} 
\caption{Forward pass and backward pass during the training of DNN using mini-batch SGD with batch size $B$ }.
\label{fig:BackProp}
\end{figure}

In Figure \ref{fig:BackProp}, $x$ is input, $y^\prime$ is predicted output during forward pass, $l_n$ is the $n$-th layer in the network. Whereas $h_n$ and $\delta_n$ are the activation (in forward pass) and error-term (in backward pass) of $n$-th layer in network.  Let $W^{n}$ be the parameters of $n$-th layer and $J(W,b;x,y)$ be the cost function of network. $b$ is the bias term and $y$ is the actual label of input training sample. 

In forward pass, $y^\prime$ is calculated as follows:
\begin{align}
\label{eqn:FP1}
h_i &= f(h_{i-1}^TW_i)   \\
\label{eqn:FP2}
y^\prime &= softmax(h_{n-1}^TW_n)  
\end{align}

In the case of a convolutional layer, $n$ is the index of the convolutional layer, and $j$ is the filter index within a convolutional layer. During backward pass error term and filter weights are updated as follows.

\begin{align}
\label{eqn:BP1}
\delta_j^i &= \Big((W_j^i)^T \delta_j^{(i+1)} \Big)\odot f^\prime(h_j^i)  \\
\label{eqn:BP2}
W_j^i &= W_j^i - \eta h_{i-1}^T \delta_i (\text{Weight update equation})
\end{align}

During the forward pass (Equations \ref{eqn:FP1} and \ref{eqn:FP2}), weight matrix $W$ is being multiplied with activations; whereas, during the backward pass, the transpose of weight matrix $(W_j^i)^T$ is being multiplied with the error-term (Equation \ref{eqn:BP1}). Therefore, the effect of {\tt Gemv2T} and {\tt Gemv2N} kernels get inter-changed during backward pass. That is, {\em in forward pass higher percentage of  {\tt Gemv2N} kernel is the main source for SM under-utilization, while, in backward pass, it is a higher percentage of  {\tt Gemv2N} kernel}.

As shown in Figure \ref{fig:SecureMobileNet}, there is substantial difference between layer-wise $BP_t$ and $FP_t$ in MobileNet-V1, because the contribution of both {\tt Gemv2T} and  {\tt Gemv2N} kernels are very high (Table \ref{tab:KernelResults}). In case of SqueezeNext, the higher percentage of {\tt Gemv2T}(Table \ref{tab:KernelResults}) led to higher rate of increase in $BP_t$  with $B$ compared to the $FP_t$ with higher $B$ (because there is lower percentage of {\tt Gemv2N}). Therefore, for SqueezeNext variants, $\frac{BP_t}{FP_t}$ ratio increases with batch size $B$ (Figure \ref{fig:MemfpAndFPbpRatio}(b) and \ref{fig:MemfpAndFPbpRatio}(d)).

\end{document}